\documentclass{article}

\usepackage{caption}
\usepackage{subcaption}

\usepackage[dvipsnames]{xcolor}

\usepackage{graphicx}
\usepackage{parskip}
\usepackage{hyphenat}
\usepackage{amsfonts,amsmath,amssymb}
\usepackage{hhline}
\usepackage{appendix}
\usepackage{bm}

\usepackage{bbm,enumerate,subcaption}
\usepackage[linesnumbered,ruled]{algorithm2e}
\SetKwFor{For}{for (}{) $\lbrace$}{$\rbrace$}
\let\oldnl\nl% Store \nl in \oldnl
\newcommand{\nonl}{\renewcommand{\nl}{\let\nl\oldnl}}% Remove line number for one line
\SetAlgoLined\DontPrintSemicolon

\usepackage{siunitx}
\usepackage{caption}
\usepackage{subcaption}

\usepackage{wrapfig}

\usepackage{microtype}
\usepackage{graphicx}
\usepackage{booktabs} % for professional tables
\usepackage{makecell}
\usepackage{tabularx}

\usepackage{booktabs, multirow} % for borders and merged ranges
\usepackage{soul}% for underlines
\usepackage{changepage,threeparttable} % for wide tables

\usepackage{hyperref}

\newcommand{\STAB}[1]{\begin{tabular}{@{}c@{}}#1\end{tabular}}

\usepackage[accepted]{icml2022}
\icmltitlerunning{Data-SUITE: Data-centric identification of in-distribution incongruous examples}

\usepackage{amsmath}
\usepackage{amssymb}

\usepackage{amsthm}
\theoremstyle{definition}
\newtheorem{definition}{Definition}[section]

\theoremstyle{property}
\newtheorem{property}{Property}[section]

\theoremstyle{assumption}
\newtheorem{assumption}{Assumption}[section]
\theoremstyle{remark}
\newtheorem*{remark}{\textbf{Remark}}

\newcommand{\X}{\mathcal{X}}

\renewcommand{\H}{\mathcal{H}}

\newcommand{\Dtrain}{\mathcal{D}_{\textrm{train}}}
\newcommand{\Dtest}{\mathcal{D}_{\textrm{test}}}
\newcommand{\Dtrainb}{\mathcal{D}_{\textrm{train2}}}
\newcommand{\Dcal}{\mathcal{D}_{\textrm{cal}}}

\newcommand{\D}{\mathcal{D}}
\renewcommand{\P}{\mathbb{P}}
\newcommand{\Ph}{\hat{\P}}
\newcommand{\R}{\mathbb{R}}
\newcommand{\N}{\mathbb{N}}

\newcommand{\set}[1]{\left\{ #1 \right\}}

\begin{document}

\twocolumn[
\icmltitle{Data-SUITE: Data-centric identification \\of in-distribution incongruous examples}




\begin{icmlauthorlist}
\icmlauthor{Nabeel Seedat}{cam}
\icmlauthor{Jonathan Crabbé}{cam}
\icmlauthor{Mihaela van der Schaar}{cam,at,ucla}
\end{icmlauthorlist}

\icmlaffiliation{cam}{Department of Applied Mathematics and Theoretical Physics, University of Cambridge, UK}
\icmlaffiliation{at}{The Alan Turing Institute, London, UK}
\icmlaffiliation{ucla}{University of California, Los Angeles, USA}

\icmlcorrespondingauthor{Nabeel Seedat}{ns741@cam.ac.uk}
\icmlkeywords{data-centric AI, data quality, machine learning, uncertainty}

\vskip 0.3in
]



\printAffiliationsAndNotice{} % otherwise use the standard text.

\begin{abstract}
Systematic quantification of data quality is critical for consistent model performance. Prior works have focused on out-of-distribution data. Instead, we tackle an understudied yet equally important problem of characterizing incongruous regions of in-distribution (ID) data, which may arise from feature space heterogeneity. To this end, we propose a paradigm shift with Data-SUITE: a data-centric AI framework to identify these regions, independent of a task-specific model. Data-SUITE leverages copula modeling, representation learning, and conformal prediction to build feature-wise confidence interval estimators based on a set of training instances. These estimators can be used to evaluate the congruence of test instances with respect to the training set, to answer two practically useful questions: (1) which test instances will be reliably predicted by a model trained with the training instances? and (2) can we identify incongruous regions of the feature space so that data owners understand the data’s limitations or guide future data collection?
We empirically validate Data-SUITE's performance and coverage guarantees and demonstrate on cross-site medical data, biased data, and data with concept drift, that Data-SUITE best identifies ID regions where a downstream model may be reliable (independent of said model). We also illustrate how these identified regions can provide insights into datasets and highlight their limitations.
\end{abstract}

\section{Introduction} \label{sec:introduction}
Machine learning models have a well-known reliance on training data quality \cite{park2021reliable}. Hence, when deploying such models in the real world, the reliability of predictions depends on the data's congruence with respect to the training data. Significant literature has focused on identifying data instances that lie out of the training data's distribution (OOD). This includes label shifts~\cite{ren2018learning,hsu2020generalized} or input feature shift, where these instances fall out of the support of the training set's distribution \citep{zhang2021understanding}. However, a much less studied, yet equally important problem is identifying heterogeneous regions of in-distribution (ID) data.

Data in the wild can be ID yet have heterogeneous regions in feature space. This manifests in varying levels of \emph{incongruence}, in cases of different sub-populations, data biases or temporal changes \cite{leslie2021does, gianfrancesco2018potential,  obermeyer2019dissecting}. We illustrate each of these types of incongruence with real world data (Table \ref{tab:data_summary}), in the experiments from Secs.~\ref{experiment3} and \ref{experiment4}.

In this paper, we present a \emph{data-centric} framework to characterize such \emph{incongruous} regions of \emph{ID data} and define two groups, namely (i) \emph{inconsistent} and (ii) \emph{uncertain}, with respect to the training distribution. We contextualize the difference based on confidence intervals (CI) (See Sec.\ref{subsec:inconsistent_and_uncertain} for details). When feature values lie outside of a CI, we term it inconsistent, alternatively we characterize the level of feature uncertainty based on the CI's width. 

At this point, one might ask if the data is ID; why should we worry? Not accounting for these incongruous ID regions of the feature space can be problematic when deploying models in high-stakes settings such as healthcare, where spurious model predictions can be deadly \cite{saria2019tutorial,varshney2020mismatched}. That said, even in settings where poor predictions are not risky, consistent exploratory data analysis (EDA) and retroactive auditing of such data is time-consuming for data scientists \cite{polyzotis2017data,kandel2012enterprise}. Hence, systematically identifying these incongruous regions has immense practical value.

\begin{figure*}[t]
    \centering
    \includegraphics[width=0.725\textwidth]{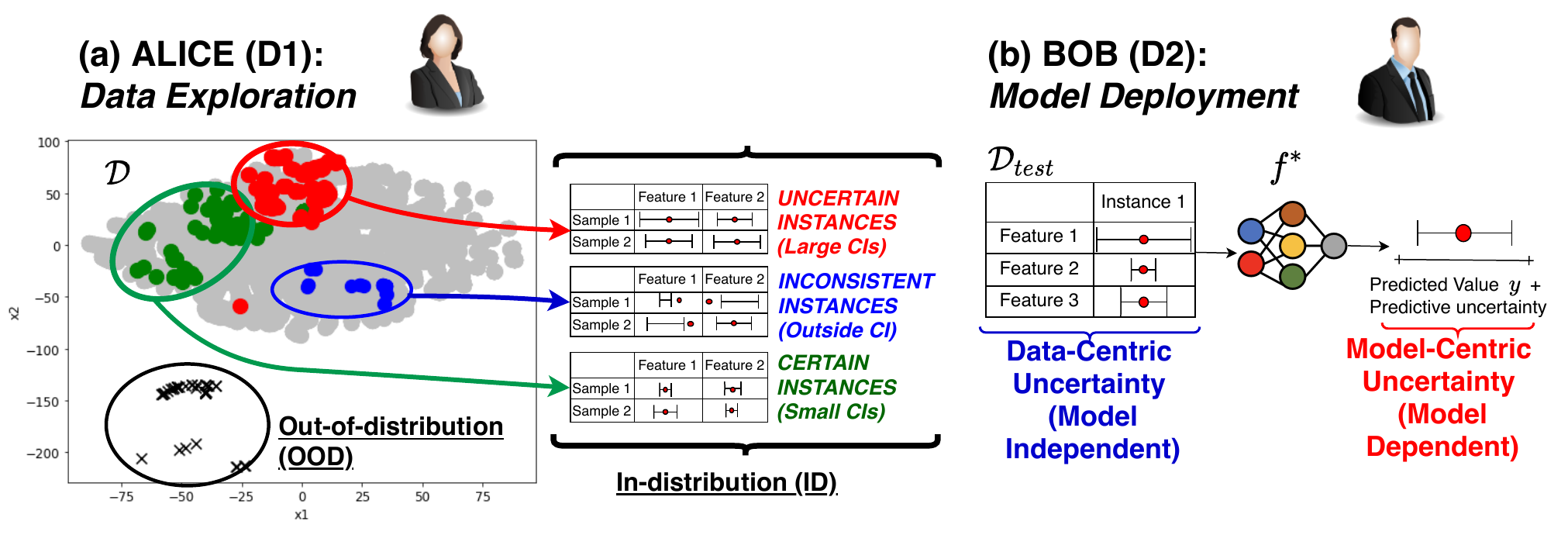}
    \vspace{-0.5cm}
    \caption{Illustration highlighting two problems Data-SUITE addresses}
    \label{fig:overview}
     \vspace{-0.2cm}
\end{figure*}

Consequently, we build a framework to empower data scientists to address the previously
mentioned challenges related to insightful exploratory data
analysis (EDA) and reliable model deployment, anchored
by the following desiderata: \\
\textbf{(D1) Insightful Data Exploration}: Alice has a new dataset $\D$ and wants to explore and gain insights into it with respect to a training set, without necessarily training a model. It would be useful if, \emph{independent of a predictive model}, she could both identify the incongruous regions of the feature space (e.g., sub-population bias or under-representation), as well as, obtain easily digestible prototype examples of each region. This could guide where to collect more data and, if this is not possible, to understand the data's limitations. \\
\textbf{(D2) Reliable Model Deployment}: Bob has a trained model $f^{*}$ and now deploys it to another site. For new data $\Dtest$, it would be useful if he could identify incongruous regions, for which he should NOT trust $f^{*}$ to make predictions. \\
\textbf{ (D3) Practitioner confidence}: Both Alice and Bob want to feel confident when using any tool. Guarantees of coverage of predictive intervals (e.g. CIs) could assist in this regard.

These examples, shown in Fig.~\ref{fig:overview} highlight the need to understand incongruence in data. As we shall discuss in the related work, there has been significant work on uncertainty estimation, with a focus on the uncertainty of a model's predictions (\emph{model-centric}). Estimating predictive uncertainty can address Bob's use-case (\emph{D2}), however since it requires a predictive model it is not naturally suited to Alice's insights use-case (\emph{D1}). Further, most predictive uncertainty methods do not provide coverage guarantees (\emph{D3}).

Therefore, in satisfying all the desiderata, we take a different approach and advocate for a \textit{data-centric} approach, where we model the uncertainty in the data features \footnote{Here, ``feature'' uncertainty refers to the degree of incongruity with the training distribution, rather than the uncertainty of the measured value (e.g. measurement noise)}. This is different from \textit{model-centric} predictive uncertainty, as we construct CIs (at feature level), without reference to any downstream model. A benefit of the flexibility is that we can flag instances and draw insights that are not model-specific (i.e. \emph{model independent}).

To ensure clarity, we note that the term \textit{data-centric} is used in the context of \textit{data-centric AI}, which we define as ``tools applied to the underlying data used to train and evaluate models''. We note that both paradigms of ML (model and data-centric) rely on models, differing based on how models are used. In ``model-centric ML'': models are used for predictive tasks and in ``data-centric ML'': models are used to study/evaluate the data itself, which does not preclude using algorithms to process the data. Our definition is consistent with \cite{polyzotis2017data} ``data-centric AI as the problem of (...) [A]nd  quality monitoring processes for datasets''. This paper fits this as a systematic tool for data quality evaluation based on uncertainty.

In this work, we focus on tabular data, a common format in medicine, finance, manufacturing etc, where data is based on relational databases  \cite{ borisov2021deep, yoon2020vime}. That said, compared to image data, tabular data has an added challenge since specific features may be uncertain while others are not; hence characterizing an instance as a whole is non-trivial. 

\vspace{-0.3cm}
\paragraph{Contributions.} We present \textbf{Data} \textbf{S}earching for \textbf{U}ncertain and \textbf{I}nconsistent \textbf{T}est \textbf{E}xamples (\textbf{Data-SUITE}), a \emph{data-centric} framework to identify incongruous regions of data using CI's and make the following contributions:\\
\textbullet\ Data-SUITE is a paradigm shift from model-centric uncertainty and, to the best of our knowledge, the first to characterize ID regions in a systematic \emph{data-centric, model-independent} manner. Not only is this more flexible, but also enables us to gain insights which are not model-specific. \\
\textbullet\ Data-SUITE's pipeline-based approach to construct feature-wise CIs enables specific properties (Sec.~\ref{feature-cis}) that permit us to flag uncertain and inconsistent instances, making it possible to identify incongruous data regions.\\
\textbullet\ Data-SUITE's performance and properties, such as coverage guarantees, are validated to satisfy \emph{D3} (Sec.~\ref{experiment1}).\\
\textbullet\  Further motivating the paradigm shift, we empirically highlight the performance benefit of a data-centric approach compared to a model-centric approach (Sec.~\ref{experiment2}). \\
\textbullet\ As a portrayal of reliable model deployment (\emph{D2}), we show on real-world datasets with different types of incongruence, that Data-SUITE best identifies incongruous data regions, translating to the best performance improvement. (Sec.~\ref{experiment3}). \\
\textbullet\ Finally, we illustrate with multiple use-cases how Data-SUITE can be used as a model-independent tool to facilitate insightful data exploration, hence satisfying \emph{D1} (Sec.~\ref{experiment4}).

\section{Related work} \label{sec:related}
This paper primarily engages with the literature on uncertainty quantification and contributes to the nascent area of data-centric AI. We also highlight the key differences of our work with the literature on noisy labels. 

\textbf{Uncertainty quantification.} 
There are numerous Bayesian and non-Bayesian methods for uncertainty quantification, including Gaussian processes \cite{williams2006gaussian}, Quantile Regression \cite{koenker2001quantile}, Bayesian Neural Networks \cite{ghosh2018structured,graves2011practical}, Deep Ensembles \cite{lakshminarayanan2017simple},  Dropout \cite{gal2016dropout, chan2020unlabelled} and Conformal Prediction \cite{vovk2005conformal}. These methods typically assess predictive uncertainty, i.e., measuring the certainty in the model's prediction \cite{seedat2019towards}. The predominant focus on predictive uncertainty is different from the notion of uncertainty in our setting, which is feature (i.e. data) uncertainty.  We specifically highlight that we quantify data uncertainty,  independent of a task-specific model. Additionally, the aforementioned methods often do not assess the coverage or provide guarantees of the uncertainty interval \cite{wasserman2004all,alaa2020discriminative} (i.e., how often the interval contains the true value). The concept of coverage will be outlined further in Secs. \ref{sec:formulation} and \ref{sec:experiments}.

\textbf{Data-Centric AI.}
Ensuring high data quality is a critical but often overlooked problem in ML, where the focus is optimizing models \cite{Sambasivan,jain2020overview}. Even when it is considered, the process of assessing datasets is adhoc or artisanal \cite{Sambasivan,ng2021}. However, there has been recent discussion around data-centric AI (DCAI), which we define as tools applied to the underlying data used to train and evaluate models, independent of the task-specific, predictive models. Our work contributes to this nascent body of work – presenting Data-SUITE, which, to the best of our knowledge, is the first systematic data-centric framework to model uncertainty in datasets. Specifically, we model the uncertainty in the feature (data) values themselves (\emph{data-centric}), which contrasts to modeling the uncertainty in predictions (\emph{model-centric}). We also highlight a tangential of classical data management \cite{kumar2017data}, which does not consider uncertainty and randomness in the data for subsequent analysis.
An exception is probabilistic databases \cite{suciu2011probabilistic}.

\textbf{Noisy labels.} Learning with noisy data is a widely studied problem, we refer the reader to \cite{algan2021image, song2020learning} for an in depth review. In machine learning, the focus is label noise.  We argue that work on noisy labels is not directly related, as the goal is to learn a model robust to the label noise, which is different from our goal of modeling the uncertainty in the features. Additionally, methods are often coupled to the task-specific predictive model, which is different from our model-independent setting.

\section{Data-SUITE} \label{sec:formulation}
In this section, we give a detailed formulation of Data-SUITE \footnote{https://github.com/seedatnabeel/Data-SUITE},\footnote{https://github.com/vanderschaarlab/mlforhealthlabpub/\newline tree/main/alg/Data-SUITE}.
We start with a problem formulation and outline the motivation for working with feature confidence intervals (CIs). Then, we describe how these CIs are built by leveraging copula modelling, representation learning and conformal prediction. Finally, we demonstrate how these CIs permit to flag uncertain and inconsistent instances.

\subsection{Preliminaries}
We consider a feature space $\X = \prod_{i=1}^{d_X} [a_i, b_i] \subseteq \R^{d_X} $, where $[a_i, b_i]$ is the range for feature $i$. Note that we make the range of each feature explicit, this will be necessary in the definition of our formalism. We assume that we have a set of $M \in \N^*$ \emph{training instances} $\Dtrain =\set{x^m \mid m \in [M]}$ sampled from an unknown distribution $\P$, where $[M]$ denotes the positive integers between $1$ and $M$. These instances typically correspond to training data for a model on a downstream task, such as classification. 

We assume that we are given new test instances $\Dtest$. Our purpose is to flag the subset of instances from $\Dtest$ that are quantitatively different from instances of $\Dtrain$ without necessarily being OOD. To that aim, we use $\Dtrain$ to build CIs $[l_i(x), r_i(x)] \subseteq [a_i, b_i]$ for each feature $i \in [d_X]$ of each test instance $x \in \Dtest$. As we will show in Sec.~\ref{subsec:inconsistent_and_uncertain}, these CIs permit to systematically flag test instances whose features are uncertain or inconsistent with respect to $\Dtrain$. For now, let us motivate the usage of feature CIs: (1) With a model of uncertainty and inconsistency at the feature level, it is possible to identify regions of the feature space $\X$ where bias and/or low coverage occurs with the training data $\Dtrain$. (2) Since CIs are built with $\Dtrain$ and without reference to any predictive downstream model, the flagged instances in $\Dtest$ are likely to be problematic for any downstream model trained on top of $\Dtrain$. Hence, we are able to draw conclusions that are not model-specific. These two points are illustrated in our experiments from Sec.~\ref{sec:experiments}. Let us now detail how the CIs are built.  

\subsection{Feature CIs}\label{feature-cis}

We now build CIs $[l_i(x), r_i(x)] \subseteq [a_i, b_i]$ for each feature $i \in [d_X]$ of each test instance $x \in \Dtest$. It goes without saying that the CIs should satisfy some properties, i.e.\\
\emph{(P1) Coverage:} We would like to guarantee that the feature $x_i$ of an instance $x \sim \P$ lies within the interval such that $\mathbb{E}\left[1_{x_{i} \in\left[l_i(x), r_i(x) \right]}\right]\geq 1-\alpha$ where the significance level $\alpha \in (0,1)$ can be chosen. In this way, a feature out of the CI hints that $x$ is unlikely to be sampled from $\P$ at the given significance level. This is then considered across all features to characterize the instance (see Sec. \ref{subsec:inconsistent_and_uncertain}).\\

\emph{(P2) Instance-wise:} The CI should be \emph{adaptive at an instance level}. i.e, we do not wish $r_i(x) - l_i(x)$ to be constant w.r.t $x \in \X$. In this way, the CIs permit to order various test instances $x \in \Dtest$ according to their uncertainty. This property is particularly desirable in healthcare settings where we wish to quantify variable uncertainty for individual patients, rather than for the population as whole.\\

\emph{(P3) Feature-wise:} We build CIs $[l_i(x), r_i(x)]$ for each feature $i \in [d_X]$ as opposed to an overall confidence region $R(x) \subset \X$. While less general than the latter approach, feature-wise CIs are more interpretable, allowing attribution of inconsistencies and uncertainty to individual features.\\

\emph{(P4) Downstream coupling:} Instances with smaller CIs are more reliably predicted by a downstream model trained on $\Dtrain$. More precisely, our CIs should have a negative correlation between CI width and downstream model performance. In this way, CIs allow to draw conclusions about the incongruence of test instances $x \in \Dtest$. 

\begin{figure}[!t]
    \centering
    \includegraphics[width=\linewidth]{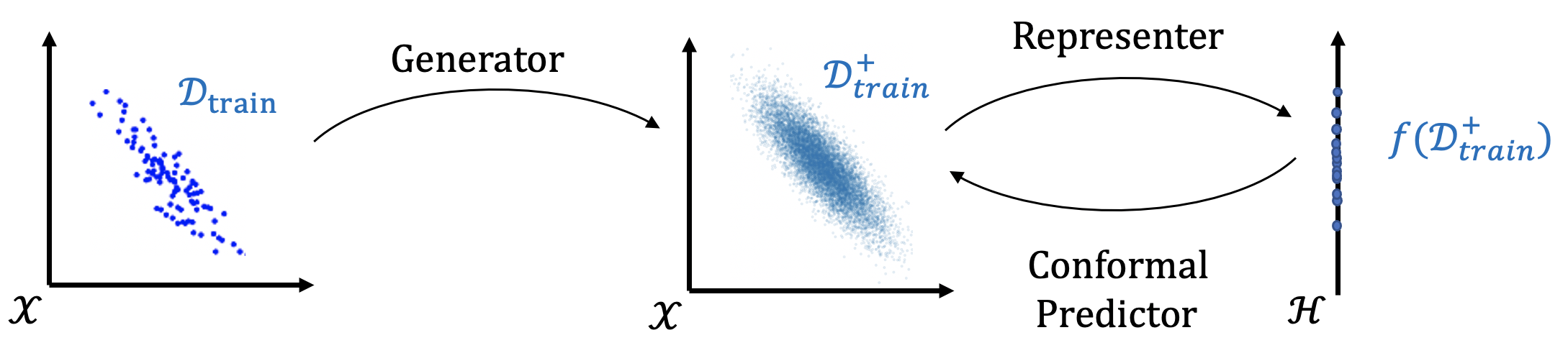}
    \vspace{-0.8cm}
    \caption{Outline of our framework \textbf{Data-SUITE}.}
    \label{fig:method}
    \vspace{-0.75cm}
\end{figure}

To construct feature CIs that satisfy these properties, we introduce a new framework leveraging copula modeling, representation learning and conformal prediction. The blueprint of our method is presented in Fig.~\ref{fig:method}. Concretely, our method relies on 3 building blocks: a \emph{generator} that augments the initial training set $\Dtrain$; a \emph{representer} that leverages the augmented training set $\Dtrain^+$ to learn a low-dimensional representation $f : \X \rightarrow \H$ of the data and a \emph{conformal predictor} that predicts instance-wise feature CIs $[l_i(x), r_i(x)]$ on the basis of each instance's representation $f(x) \in \H$. By construction of the CIs, this method fulfills properties (P2) and (P3). As we will see in the following, the conformal predictor's theoretical properties guarantees (P1). However, we also empirically validate (P1), see Sec \ref{experiment1}.  We then also demonstrate (P4) empirically in Sec.~\ref{sec:experiments}. Appendix \ref{ablation} quantifies the significance of each block via an ablation study. Let us now detail each block. 

\paragraph{Generator.} The purpose of the generator is to augment the initial training set $\Dtrain$ with instances that are consistent with the initial distribution $\P$. Many data augmentation techniques can be used for this block. Since our focus is on tabular data, we found \emph{copula modeling} to be particularly useful. Copulas leverage Sklar's theorem~\cite{Skla59} to estimate multivariate distributions with univariate marginal distributions. In our case, we use vine copulas~\cite{bedford2001probability} to build an estimate $\Ph$ for the distribution $\P$ on the basis of $\Dtrain$. We then build an augmented training set $\Dtrain^+$ by sampling from the copula density $\Ph$. Interestingly, our method does not need to access $\Dtrain$ once the copula density $\Ph$ is available. It is perfectly possible to use only instances from $\Ph$ to build the augmented dataset $\Dtrain^+$. This could be useful for data sharing, if the access to the training set $\Dtrain$ is restricted to the user. Further details and motivations on copulas is found in Appendix~\ref{appendix:copula}. Note that a copula might not be ideal for very high-dimensional (large $d_X$) data in domains such as computer vision or genomics. In those cases, copula modeling can be replaced by domain-specific augmentation techniques.  

\paragraph{Representer.} A trivial way to verify the coverage guarantee (P1) would be to use the true values of the features to build the CIs: $[l_i(x), r_i(x)] = [x_i - \delta, x_i + \delta]$ for some $\delta \in \R^+$. The problem with this approach is two-fold: (1) it does not leverage the distribution $\P$ underlying the training set $\Dtrain$ and (2) it results in an uninformative reconstruction with CIs that does not capture the specificity of each instance, hence  contradicting (P2). To provide a more satisfactory solution, we propose to represent the augmented training data $\Dtrain^+$ with a representation function $f : \X \rightarrow \H$ that maps the data into a lower-dimensional latent representation space $\H \subseteq \R^{d_H}$, $d_H < d_X$. The purpose of this representer is to capture the structure of the low-dimensional manifold underlying $\Dtrain^+$.
At test time, the conformal predictor (detailed next), uses the lower representations $f(x) \in \H$ to estimate a reconstruction interval for each feature $x_i$. This permits to bring a satisfactory solution to the two aforementioned problems: (1) the CIs are reconstructed in terms of latent factors that are useful to describe the training set $\Dtrain$ and (2) the predicted CIs vary according to the representation $f(x) \in \H$ of each test instance $x \in \Dtest$. In essence, our approach is analogous to autoencoders. As we will explain soon, the crucial difference is the decoding step: our method outputs CIs for the reconstructed input. In this work, we use Principal Component Analysis (PCA), the workhorse for tabular data, to learn the representer $f$. Note that more general encoder architectures can be used in settings such as computer vision. 
\vspace{-0.3cm}

\begin{figure}[!t]
    \centering
    \includegraphics[width=.75\linewidth]{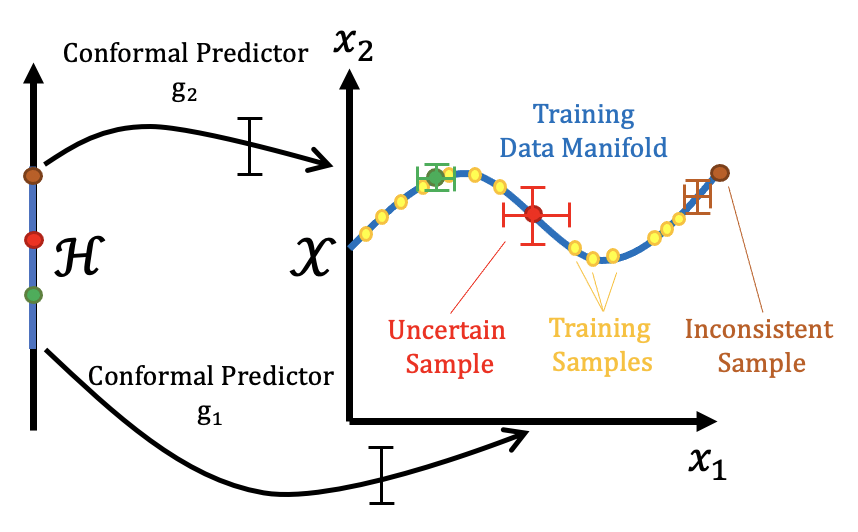}
    \caption{Conformal Predictor in \textbf{Data-SUITE}.}
    \label{fig:conformal_illustration}
    \vspace{-0.75cm}
\end{figure}

\paragraph{Conformal Predictor.} We now turn to the core of the problem: estimating feature-wise CIs. As previously mentioned, the CIs $[l_i(x), r_i(x)], i \in [d_X]$ will be \emph{computed on the basis of the latent representation $f(x)$} for each $x \in \Dtest$. The idea is simple: for each feature $i \in [d_X]$, we train a regressor $g_i : \H \rightarrow [a_i, b_i]$ to reconstruct an estimate of the initial features $x_i$ from the latent representation $f(x)$ of the associated training instance: $(g_i \circ f)(x) \approx x_i$. We stress that the regressor $g_i$ has no knowledge of the true observed $x_i$ but only of the latent representation $f(x)$, as illustrated in Fig.~\ref{fig:conformal_illustration}. Of course, the feature regressor by themselves provides point-wise estimates for the features. In order to turn these into CIs, we use conformal prediction as a wrapper around the feature regressor~\cite{vovk2005conformal}. 

We formalize our problem in the framework of Inductive Conformal Prediction (for motivations, see Appendix \ref{appendix:conformal}). Hence, under the formulation, we start by splitting the augmented training set into a proper training set and a calibration set: $\Dtrain^+ = \Dtrainb^+ \sqcup \Dcal^+$. We use the latent representation of the proper training set, to train the feature regressor $g_i , i \in [d_X]$ for the reconstruction task. Then, the latent representation of the calibration set is used to compute the non-conformity score ($\mu$), which estimates how different a new instance looks from other instances. 

In practice, we use the absolute error non-conformity score $\mu_i(x) =  |x_i - (g_i \circ f)(x)|$. We obtain an empirical distribution of non-conformity scores $\{ \mu_i(x) \mid x \in \Dcal^{+} \}$ over the calibration instances for each feature $i \in [d_X]$. This is used to obtain the critical non-conformity score $\epsilon$, which corresponds to the $\lceil(|\Dcal^+| + 1)(1-\alpha)\rceil$-th smallest residual from the set $\{ \mu_i(x) \mid x \in \Dcal^+ \}$  \cite{vovk2013transductive}.
We then apply the method to any unseen incoming data to obtain predictive CIs for the data point, i.e. $[l_i(x), r_i(x)] = [(g_i \circ f)(x)-\epsilon,  (g_i \circ f)(x)+\epsilon]$. However, in this form the CIs are constant for all instances, where the width of the interval is determined by the residuals of the most difficult instances (largest residuals).

We adapt our conformal prediction framework to obtain the desired adaptive intervals (\emph{P2}) using a normalized non-conformity function ($\gamma$), see Eq. \ref{norm_conformity} \cite{bostrom2016evaluation,Johansson}. The numerator is computed as before based on $\mu$, however, the denominator normalizes per instance. We learn the normalizer per feature $i \in [d_X]$. 

To do so, we compute the log residuals per feature, for all instances in the respective proper training set $\Dtrainb$. We produce tuples per feature: $\{(f(x), ln|x - (g_i \circ f)(x)|) \mid x \in \Dtrainb \}$. These are used to train a \emph{different} model, $\sigma_i : \X \rightarrow \R^+$ (e.g., MLP), to predict the log residuals.  We can then apply $\sigma_i$ to \emph{test instances} to capture the difficulty in predicting said instance. Note, we apply an exponential to the predicted log residual for the test instance converting to the true scale and ensuring positive estimates.
\begin{align}\label{norm_conformity}
    \gamma_i(x) \equiv \frac{|x_i - (g_i \circ f)(x)|}{\sigma_i(x)},
\end{align}
We can then obtain the critical non-conformity score $\epsilon$ applied to the empirical distribution of normalized non-conformity scores $\{\gamma_i(x) \mid x \in \Dcal^+ \}$, in the same way as before, based on residuals. 

The instance-specific adaptive intervals are then obtained as per Eq. \ref{instance_pred}, where $g$ is the underlying feature regressor and $\sigma_i$ is the instance-wise normalizing function. 

\begin{align}\label{instance_pred}
      [l_i(x), r_i(x)] = [(g_i \circ f)(x)-\epsilon  \sigma_i(x), (g_i \circ f)(x) + \epsilon  \sigma_i(x)]
\end{align}
\paragraph{Remarks on theoretical guarantees.} Under the exchangeability assumption detailed in the Appendix \ref{appendix:conformal}, the validity of coverage guarantees (P2) is fulfilled with our definition. In our implementation, we use $\alpha = .05$.

\subsection{Identifying Inconsistent and Uncertain Instances} \label{subsec:inconsistent_and_uncertain}
Now that we have CIs $[l_i(x), r_i(x)] \subset [a_i, b_i]$ for each feature $x_i , i \in [d_X]$ of the instance $x$, we can evaluate if instances from a dataset falls within the predicted range. If it falls outside the predicted range, we characterize the inconsistency (see Definition \ref{def:incons})

\begin{definition}[Inconsistency] 
Let $x \in \Dtest$ be a test instance for which we construct a $(1-\alpha)$-CI, $[l_i(x), r_i(x)]$, for each feature $x_i , i \in [d_X]$ for some predetermined $\alpha \in (0,1)$. For each $x_i , i \in [d_X]$, the \emph{feature inconsistency} is a binary variable indicating if $x_i$ falls out of the CI. 
\begin{align}
     \nu_i(x) \equiv  1 (x_{i} \notin [l_{i}(x), r_{i}(x)])
\end{align}
The \emph{instance inconsistency} $\nu(x)$ is obtained by averaging over the feature inconsistencies $\nu_i(x)$.
\begin{align*}
    \nu(x) \equiv \frac{1}{d_X} \sum_{i=1}^{d_X} \nu_i (x) 
\end{align*}
The instance $x$ is \emph{inconsistent} if the fraction of inconsistent features is above a predetermined threshold\footnote{In our implementation, we use $\lambda=0.5$.} $\lambda \in [0,1]$: $\nu(x) > \lambda$.  
\label{def:incons}
\end{definition}

There can also be degrees of uncertainty in the feature value for features that fall within the CI, which can reflect the instance as a whole. Indeed, if the CI $[l_i(x), r_i(x)]$ is large, the feature $x_i$ is likely to fall within its range. Nonetheless, we should keep in mind that large CIs correspond to a large uncertainty for the related feature. This will also typically happen when the instance $x \in \Dtest$ differs from the training set $\Dtrain$ used to build the CI. We now introduce a quantitative measure that expresses the degree of uncertainty of the instance with respect to $\Dtrain$ (see Definition \ref{def:uncertainty}).

\begin{definition}[Uncertainty] 
Let $x \in \Dtest$ be a test instance for which we construct a $(1-\alpha)$ CI $[l_i(x), r_i(x)]$ for each feature $x_i , i \in [d_X]$ for a predetermined $\alpha \in (0,1)$. For each $x_i , i \in [d_X]$, we define \emph{feature uncertainty} $\Delta_i(x)$ as the feature CI width normalized by the feature range:
\begin{align}
    \Delta_i(x) \equiv  \frac{r_i(x) - l_i(x)}{b_i - a_i}
\end{align}
The \emph{instance uncertainty} $\Delta(x)$ is obtained by averaging over all feature uncertainties:
\begin{align*}
    \Delta(x) \equiv \frac{1}{d_X} \sum_{i=1}^{d_X} \Delta_i (x) \hspace{.2cm} \in (0,1].
\end{align*}
\vspace{-0.5cm}
\label{def:uncertainty}
\end{definition}
\begin{remark}

Instance uncertainties are strictly larger than zero, as feature uncertainties are computed over all features. Hence, this characterization offers a natural split between \emph{certain} and \emph{uncertain} instances if we sort the instances based on uncertainty. 
\end{remark}

%\section{Method} \label{sec:method}
%\input{parts/method}

\section{Experiments} \label{sec:experiments}
This section presents a detailed empirical evaluation demonstrating that Data-SUITE satisfies (\textbf{D1}) Insightful Data Exploration,(\textbf{D2}) Reliable Model Deployment and (\textbf{D3}) Practitioner confidence, introduced in Sec. 1. We tackle these in reverse order, as practitioner confidence is a prerequisite for the adoption of \textbf{D1} and \textbf{D2}. Recall that the notion of uncertainty in Data-SUITE is different from predictive uncertainty (model-centric). We empirically compare these two paradigms using methods for predictive uncertainty. That said, a natural additional question is whether model-centric uncertainty estimation methods can simply be applied in this setting and provide uncertainty estimates for feature values. We benchmark the following widely used Bayesian and non-Bayesian methods (under BOTH model-centric \& data-centric paradigms): Bayesian Neural Networks (BNN) \cite{ghosh2018structured}, Deep Ensembles (ENS) \cite{lakshminarayanan2017simple}, Gaussian Processes (GP) \cite{williams2006gaussian}, Monte-Carlo Dropout (MCD) \cite{gal2016dropout} and Quantile Regression (QR) \cite{koenker2001quantile}. We also ablate and test Data-SUITE's constituent components independently: conditional sampling from copula (COP), Conformal Prediction on raw data (CONF) \cite{vovk2005conformal, balasubramanian2014conformal}. For implementation details, see Appendix \ref{appendix:benchmarks}.

\subsection{Validating coverage \& comparing properties}\label{experiment1}
We firstly wish to validate the CIs to ensure that the coverage guarantees are satisfied such that users can have confidence that the true value lies within the predicted CIs (\textbf{D3}). We assess the CIs based on the following metrics defined in  \cite{navratil2020uncertainty} -  (1) Coverage: how often the CI contains the true value, (2) Deficit: extent of CI shortfall (i.e., the severity of the errors) and (3) Excess: extent of CI excess width to capture the true value.
\begin{align}
&\text {Coverage} =  \mathbb{E}\left[1_{x_{i} \in\left[l_{i}, r_{i} \right]}\right] \\
&\text {Deficit} =  \mathbb{E}\left[1_{x_i \notin[l_{i}, r_{i}]} \cdot \min \left\{\left|x_i-l_{i}\right|,\left|x_i-r_{i}\right|\right\}\right] \\
&\text {Excess} =  \mathbb{E}\left[1_{x_i \in [l_{i}, r_{i}]} \cdot \min \left\{x_i-l_{i}, r_{i}-x_i\right\}\right]
\end{align}

\textbf{Synthetic data.} 
We assess the properties of different methods using synthetic data as the ground truth values are available, even when encoding incongruence. The synthetic data with features, $\mathbf X=[X_1, X_2, X_3]$, is drawn IID from a multivariate Gaussian distribution, parameterized by mean vector $\mu$ and a positive definite covariance matrix $\sum$ (details in Appendix \ref{appendix:synthetic}). We sample $n=1000$ points for both $\Dtrain^{synth}$ and $\Dtest^{synth}$ and encode incongruence into $\Dtest^{synth}$ using a multivariate additive  model $\hat{\mathbf X} = \mathbf X+ \mathbf Z$, where $\mathbf Z \in R^{n \times m}$, is the perturbation matrix. We conduct experiments with different configurations: (1) \emph{$D_a$}: Multivariate Gaussian with variance 2 and varying proportion of perturbed instances. (2) \emph{$D_b$}: Multivariate Gaussian with varying variance and fixed proportion of perturbed instances ($50 \%$) and (3) \emph{$D_c$}: Varying distribution $\in \{Beta, Gamma, Normal, Weibull\}$.

\begin{figure}
\captionsetup[subfigure]{labelformat=empty, justification=centering}
\begin{subfigure}{0.5\textwidth}
\includegraphics[width=1\textwidth]{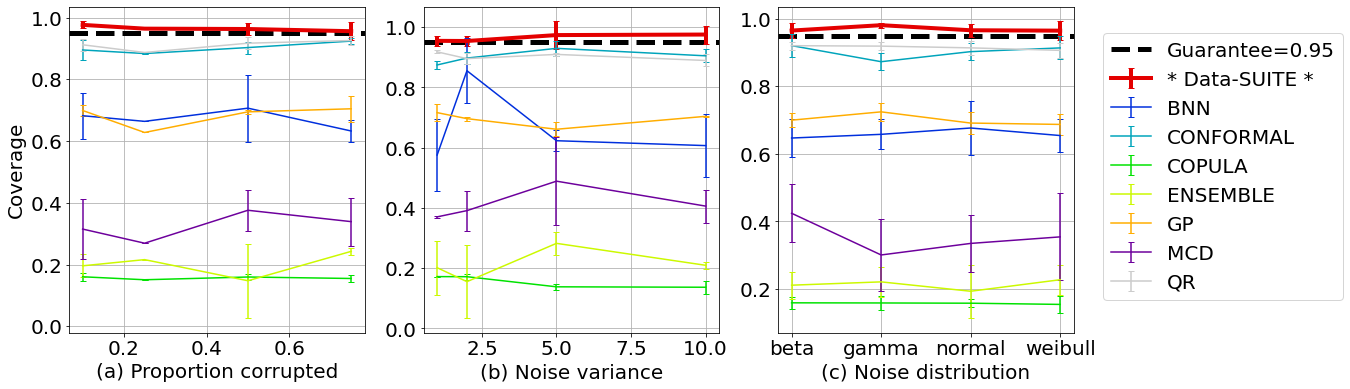}
\caption{i. Coverage \& 95\% guarantee (1-$\alpha$)}
\end{subfigure}%
\\
\begin{subfigure}{0.5\textwidth}

\includegraphics[width=1\textwidth]{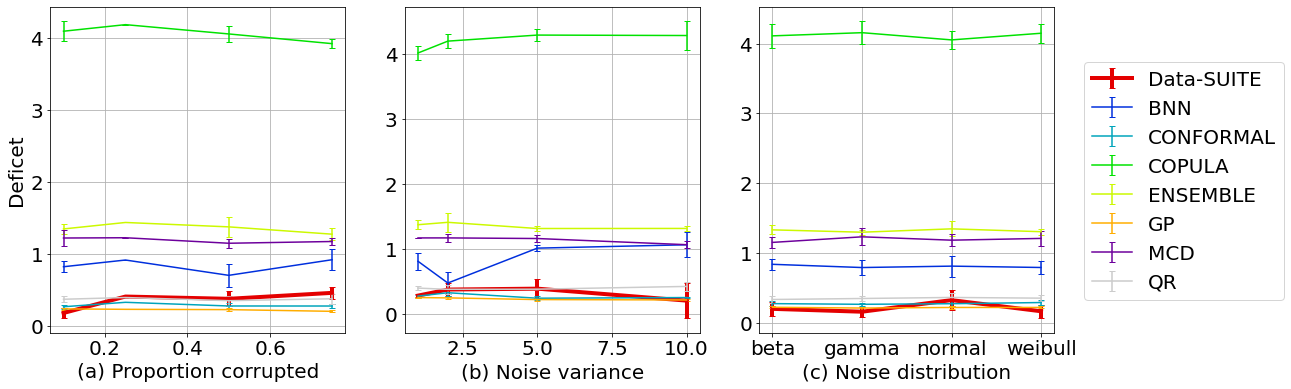}
\caption{ii.Deficit}
\end{subfigure}%
\\
\begin{subfigure}{0.5\textwidth}

\includegraphics[width=1\textwidth]{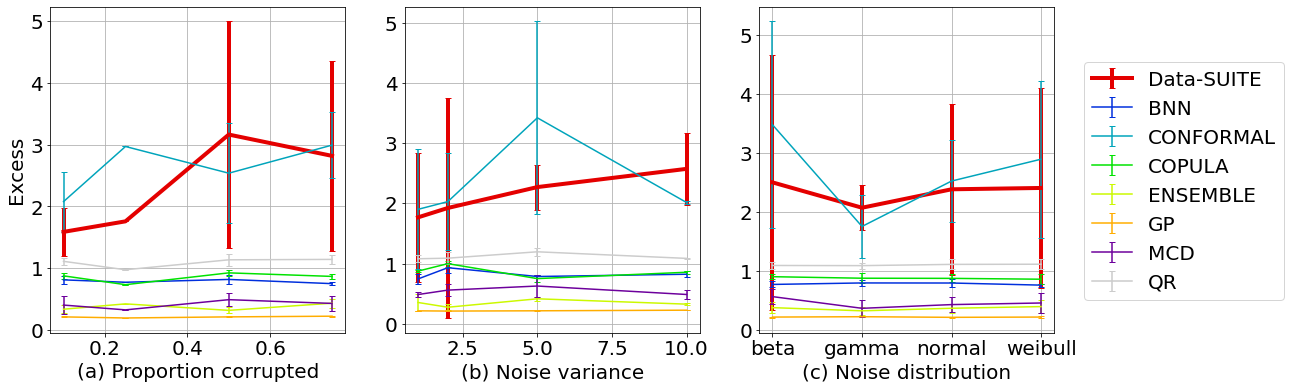}
\caption{iii. Excess}
\end{subfigure}%
\vspace{-.1cm}
\caption{Comparison of methods based on coverage, deficit and excess under various configurations ($D_a,D_b ,D_c$)}
\label{fig:synthetic}
\vspace{-.5cm}
\end{figure}

Fig.~\ref{fig:synthetic} outlines mean coverage, deficit and excess averaged over five runs for $\Dtest^{synth}$ under the different configurations ($D_a,D_b ,D_c$). There is a clear variability amongst the different methods, suggesting specific methods are more suitable. Data-SUITE outperforms the other methods based on coverage and deficit across all configurations. We note the methods with poor coverage are typically ``incorrectly'' confident, i.e. small intervals with low coverage and high deficit.
Fig.~\ref{fig:synthetic} also demonstrates a meaningful relationship that coverage and deficit are inversely related (high coverage is associated with the low deficit), as with deficit and excess. Although high coverage and low deficit ideally occur with low excess, we observe that high levels of coverage occur in conjunction with high levels of excess. Critically, however, in satisfying \textbf{D3}, Data-SUITE maintains the 95\% coverage guarantees across all configurations, unlike other methods. 

\subsection{Synthetic data stratification w/ downstream task}\label{experiment2}
While it is essential to validate a method's properties, the most useful goal is whether the intervals can be used to identify instances that will be reliably predicted by a downstream predictive model. With Data-SUITE, we stress that this is done in a model-independent manner (i.e. no knowledge of the downstream model). 
We train a downstream regression model using $\Dtrain^{synth}$, where features $X_1, X_2$ are used to predict $X_3$. We first compute a baseline mean squared error (MSE) on a held-out validation set of $\Dtrain^{synth}$ and the complete test set $\Dtest^{synth}$ ($\hat{\mathbf X} = \mathbf X+ \mathbf Z$). Thereafter, we construct predictive intervals for $\Dtest^{synth}$ using all benchmark methods (either uncertainty intervals or CIs). The intervals are then used to sort instances based on width. 

In addition, we answer the question of whether a \emph{data-centric} or \emph{model-centric} approach yields the best performance. For data-centric paradigm, we construct intervals for the features $X_1, X_2$, hence instances are categorized in a model-independent manner based on data-level CIs. In contrast, the model-centric paradigm is tightly coupled with a task-specific model, categorizing instances using predictive uncertainty based on prediction $X_3$. 

We then compute MSE for the 100 most certain instances as ranked by each method (smallest widths). For Data-SUITE, we also compute the MSE for those instances identified as \emph{inconsistent} (outside CIs). The best method is one in which the \emph{certain} sorted instances produce MSE values closest to the clean train MSE (baseline) i.e. has the lowest MSE.

Table 1 shows the MSE for configurations $D_a$ and $D_b$. As one example of satisfying \textbf{D2}, Data-SUITE has the best performance and identifies the top 100 \emph{certain} instances that yields the best downstream model performance, with the lowest MSE across all configurations. In addition, as expected the \emph{inconsistent} instances are unreliably predicted. The poor performance for ablations of Data-SUITE components, suggests the necessity of the inter-connected framework (more in Appendix \ref{ablation})

Additionally, we see for the same base methods (e.g. BNN, MCD etc), that the data-centric paradigm outperforms the model-centric paradigm in identifying the ``best'' instances to give the lowest MSE. This result highlights the performance advantage of a flexible, model-independent data-centric paradigm compared to the model-centric paradigm.

\subsection{Real dataset stratification w/ downstream task}\label{experiment3}

We now demonstrate how Data-SUITE can be practically used on real data to stratify instances for improved downstream performance (satisfying \textbf{D2}). Specifically, to assist with more reliable and performant model deployment across a variety of scenarios. To this end, we select three real-world datasets with different types of incongruence as presented in Table \ref{tab:data_summary}. For details see Appendix \ref{appendix:real}.

\begin{table}[t]
\centering
\caption{MSE based on instance stratification for different methods. Data-SUITE outperforms other methods, whilst data-centric methods in general outperform model-centric methods}
  \scalebox{.68}{
  \begin{tabular}{cl||ccc||cc}
    \hline
     & \multicolumn{1}{c||}{} & \multicolumn{3}{c||}{Proportion ($D_a$)} & \multicolumn{2}{c}{ Variance  ($D_b$)} \\ \hline
     \multirow{3}{*}{\STAB{\rotatebox[origin=c]{90}{}}}
      & PERTURBATION                     &  .1      &  .25     &   .5   & 1  &    2                           \\ \hline
     & \textit{Train Data (BASELINE)}         &  \textit{.067}      &  \textit{.059}     &   \textit{.068}   &  \textit{.065}  &    \textit{.068}                   \\
     & Test Data            &     .222   &    .513   &   .889   &   .275      &   .889               \\ \hline\hline
     \multirow{9}{*}{\STAB{\rotatebox[origin=c]{90}{\textbf{Data-centric}}}}
     &  \textcolor{ForestGreen}{\textbf{Data-SUITE (All, Uncertainty)}}  & \textcolor{ForestGreen}{\textbf{.069}}  & \textcolor{ForestGreen}{\textbf{.122}} &  \textcolor{ForestGreen}{\textbf{.197}}  &  \textcolor{ForestGreen}{\textbf{.104}} & \textcolor{ForestGreen}{\textbf{.197}}   \\
      & \textcolor{red}{Data-SUITE (All, Inconsistent)}     &     \textcolor{red}{.595}   &  \textcolor{red}{1.608}     &   \textcolor{red}{2.322}   &   \textcolor{red}{ .791}     &    \textcolor{red}{2.322}              \\ 
     & Data-SUITE (CONF) &  .125 & .396 &  .846  & .293  &  .846  \\
     & Data-SUITE (COP)   &  .220 & .277 &  .451  & .236  &  .451  \\
     & BNN    & .192  & .216 &  .704  &  .173 &   .704 \\
      & ENS    & .125  & .311 &  .565  &  .204 &  .565  \\
     & GP     &  .112  & .153 &   .296 &  .158 &  .296  \\
     & MCD       &  .173 & .391 & .692   & .201  & .692   \\
      & QR     & .116  & .228 &  .635  & .193  & .635    \\
     \hline\hline
     \multirow{4}{*}{\STAB{\rotatebox[origin=c]{90}{\textbf{Model-centric}}}}
     & BNN (Predictive)  & .208  & .220 &  .692  &  .195 & .692   \\
     & ENS (Predictive)   &  .143  & .226 &  .625  &  .257 & .625    \\
     &  GP (Predictive)    & .147  & .472 &  .584  &  .237 & .584    \\
     & MCD (Predictive)   &  .206 & .255 & .684   &  .213 &  .684  \\
     & QR (Predictive) &  .187 &  .477 &  .671   &  .223 &  .671  \\
     \hline\hline
  \end{tabular}}
  \vspace{-0.8cm}
\end{table}

\textbf{Evaluation.} We stratify $\Dtest$ into \emph{certain} and \emph{uncertain} instances based on the interval width predicted by each method. e.g. the most uncertain has the largest width.

\textbf{\textit{Are the identified instances OOD?}} At this point, one might be tempted to assume that the identified instances are simply OOD. i.e. samples that fall out of the support of the training set's distribution. We show in reality that this is unlikely the case. We apply existing algorithms to detect OOD and outliers (for others see \cite{yang2021generalized}): Mahalanobis distance \cite{lee2018simple}, SUOD \cite{zhao2021suod}, COPOD \cite{li2020copod} and Isolation Forest \cite{liu2012isolation}. For each of the detection methods, we compute the overlap between the predicted OOD/Outlier instances and the uncertain and inconsistent instances as identified by Data-SUITE. We found minimal overlap across methods, ranging between 1-18$\%$. Additionally, the OOD detection methods were often unconfident in their predictions, with average confidence scores ranging between 5-50$\%$. Both results suggest the identified uncertain and inconsistent instances are unlikely OOD. For more see Appendix \ref{appendix:ood}.

\begin{table}[t]
\centering
\caption{Comparison of real-world datasets}
\scalebox{0.68}{
\begin{tabular}{@{}ccccccc@{}}
\toprule
Dataset &  Incongruence Type     & Downstream Task & Stratification    \\
\midrule
\makecell{\textbf{Seer (US)} \& \\ \textbf{Cutract (UK)}} &  \makecell{Geographic UK-US \\(Cross-site medical)} & \makecell{Predict mortality \\from prostate cancer} & \makecell{$\Dtrain^{Seer}$: Seer (US), \\$\Dtest^{Cut}$: Cutract (UK)}  \\ \hline
\textbf{Adult} & \makecell{Demographic Bias\\ (Gender \& Income)}  & \makecell{Predict income \\over \$50K} & \makecell{$\Dtrain^{Adult}$, $\Dtest^{Adult}$ \\balanced split}  \\ \hline
\textbf{Electricity} & \makecell{Temporal \\(Consumption patterns)} & \makecell{Predict electricity \\ price rise/fall} & \makecell{$\Dtrain^{Elec}$ :1996 , \\$\Dtest^{Elec}$ : 1997-1998}\\
\bottomrule
\end{tabular}}
\label{tab:data_summary}
\vspace{-0.3cm}
\end{table}

Each stratification method will identify different instances for each group, hence we aim to quantify which method identifies instances that provide the most improvement to downstream performance. We do this by computing the accuracy of the \emph{certain} and \emph{uncertain} stratification's, on a downstream random forest trained on $\Dtrain$. Ideally, correct instance stratification results in greater accuracy for \emph{certain} compared to \emph{uncertain} instances. 
As an overall comparative metric, we compute the \emph{Mean Performance Improvement - $MPI$} (Eq. \ref{eq:perf_improvement}). MPI is the difference in accuracy ($Acc$) between \emph{certain} and \emph{uncertain} instances, as identified by a specific method, averaged over different threshold proportions $P$. The best performing method would clearly identify the most appropriate \emph{certain} and \emph{uncertain} instances, which would translate to the largest MPI.
\vspace{-0.1cm}
\begin{align}
    MPI = \frac{1}{|P|} \sum_{p \in P} Acc(Cert_{p})-Acc(Uncert_{p})
    \label{eq:perf_improvement}
\end{align}
where P = $\{0.05k ~|~ k \in$ [20] \}, $Cert_p$ = Set of $p$ most \emph{certain} instances, $Uncert_p$ = Set of $p$ most \emph{uncertain} instances.

\begin{figure}[!h]
    \centering
    \includegraphics[width=0.47\textwidth]{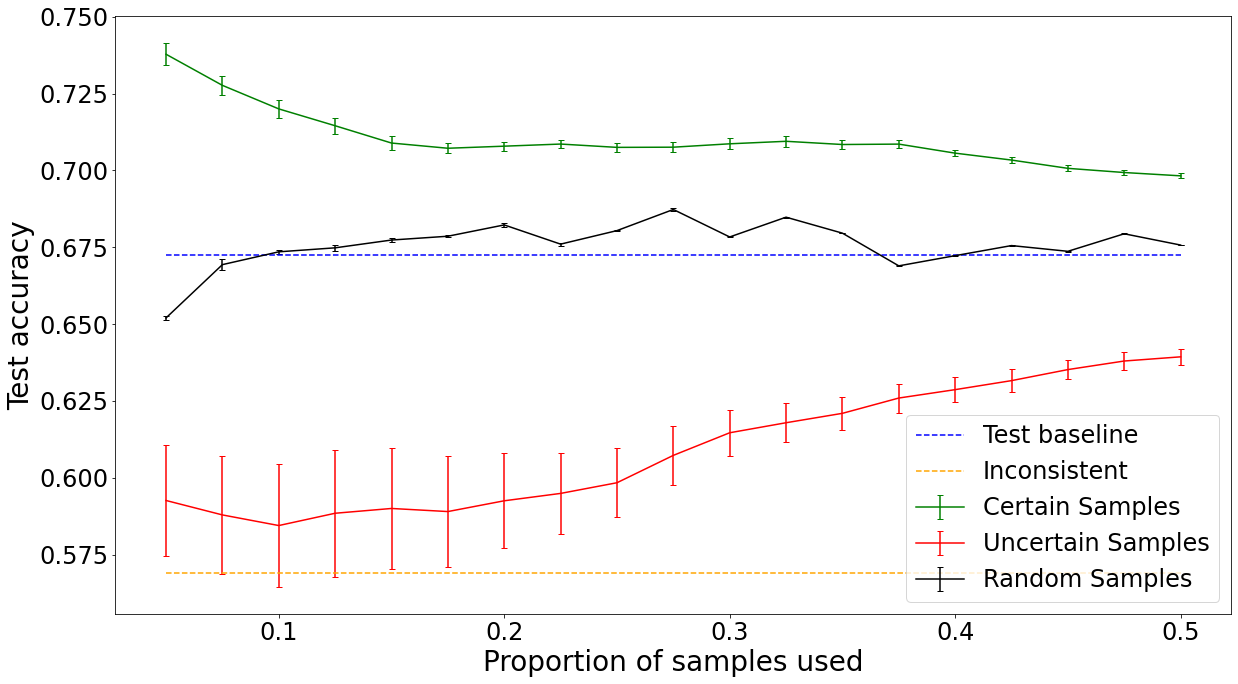}
    \caption{Example on CUTRACT of how Data-SUITE instance stratification can be used to improve downstream performance, contrasted with baseline $\Dtest$ (blue) or random selection (black). }
    \label{fig:seer_cutract}
    \vspace{-0.3cm}
\end{figure}

Fig. \ref{fig:seer_cutract} illustrates an example of Data-SUITE, applied to the CUTRACT dataset.  The metric \emph{MPI} (Eq. \ref{eq:perf_improvement}) is the mean difference between certain (green) and uncertain (red) curves. The results demonstrate the performance improvement when evaluating with the stratified \emph{certain} and \emph{uncertain} instances (compared to performance evaluated on the baseline $\Dtest$ or random sampling of instances). The result further demonstrates that the identified \emph{inconsistent} instances have worse performance when compared to \emph{uncertain} instances. 

Table \ref{tab:downstream_real} shows the \emph{MPI} scores across methods. In satisfying \textbf{D2}, of improving deployed model performance, Data-SUITE consistently outperforms other methods, providing the greatest performance improvement, with the lowest variability across datasets. The result suggests that Data-SUITE identifies the most appropriate \emph{certain} and \emph{uncertain} instances, accounting for the performance improvement. Overall, the quality of stratification by Data-SUITE has not been matched by any benchmark uncertainty estimation method.

\begin{table}[t]
    \centering
        \caption{MPI metric across datasets for different methods}
    \scalebox{0.775}{
    \begin{tabular}{lccc}
\toprule
 &      SEER-CUTRACT & Adult & Electricity    \\
\midrule
\textbf{Data-SUITE}            & \textbf{0.11 $\pm$ 0.015}  &  \textbf{0.64 $\pm$ 0.03}  & \textbf{0.26 $\pm$ 0.03} \\
BNN &  0.08 $\pm$ 0.02  &  -0.15 $\pm$ 0.02 & -0.005 $\pm$ 0.01 \\
CONFORMAL           &  0.05 $\pm$ 0.01  &  -0.07 $\pm$ 0.07 & 0.12 $\pm$ 0.03 \\
ENSEMBLE            & 0.01 $\pm$ 0.02 &  -0.03 $\pm$ 0.02 & -0.02 $\pm$ 0.02\\
GP            &  0.05 $\pm$ 0.04  &  0.56 $\pm$ 0.02 &  0.04 $\pm$ 0.04 \\
MCD            &  0.01 $\pm$ 0.01 &   -0.16 $\pm$ 0.01 & 0.15 $\pm$ 0.03 \\
QR & -0.10 $\pm$ 0.03  &   0.12 $\pm$ 0.06 & 0.15 $\pm$ 0.06 \\
\bottomrule
\end{tabular}}
    \label{tab:downstream_real}

\end{table}

\subsection{Data-SUITE usage with diverse downstream models}\label{diverse}

Data-SUITE operates to identify instances , independent of the downstream task predictive model. Previous experiments use a fixed downstream model (e.g. RF). We now evaluate a more diverse set downstream models, i.e. RF, MLP, GBT and include a ROBUST model- Median-of-Means (MOM) estimators \cite{lecue2020robust}). For each downstream model, we compare the performance on instances identified by Data-SUITE and baselines. We rank them based on MPI (see Eq.\ref{eq:perf_improvement}). i.e larger MPI=higher rank. Ideally, the rank should be invariant across models, showing the instances identified by Data-SUITE are impactful, no matter the downstream model. 

\begin{figure}[!h]
    \centering
    \includegraphics[width=0.35\textwidth]{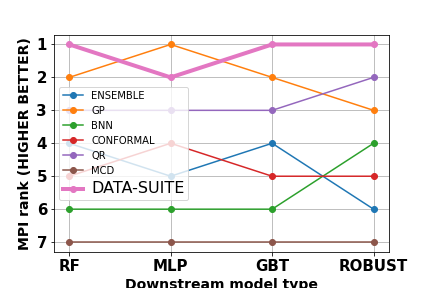}
       \vspace{-0.3cm}
    \caption{Rank assessment for a diverse set of downstream models, showing Data-SUITE has the most consistent ordering  across different models}
    \label{fig:rank}
    \vspace{-0.3cm}
\end{figure}

We show results for the Adult dataset in Fig. \ref{fig:rank}, with the other datasets included in Appendix \ref{rank_appendix}. Overall, the results are similar, whereby Data-SUITE’s identification remains the most appropriate, with consistent highest MPI rank, no matter the model. In contrast, baseline methods rank differently depending on the model. This shows that Data-SUITE identifies impactful instances no matter the downstream task-specific model.

\subsection{Use-Case: Data-SUITE in the hands of users}\label{experiment4}
We now demonstrate how users can practically leverage Data-SUITE to better understand their data. We do so by profiling the incongruous regions identified by Data-SUITE and highlight the insights which can be garnered independent of a model. This satisfies \textbf{D1}, where the quantitative profiling provides valuable insights that could assist data owners to characterize where to collect more data and if this is not possible, to understand the data's limitations. 

\begin{figure}[t]
\captionsetup[subfigure]{labelformat=empty, justification=centering}

\begin{subfigure}{0.5\textwidth}
\centering
\includegraphics[width=0.875\textwidth]{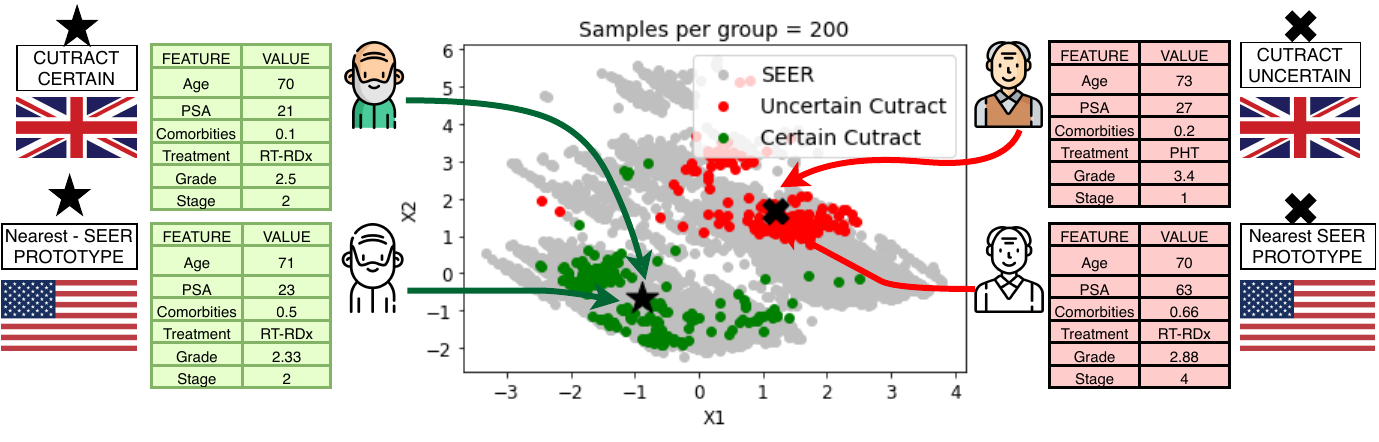}
\caption{i. SEER-CUTRACT: CUTRACT \emph{certain} instances are similar to their SEER nearest prototypes, whilst CUTRACT \emph{uncertain} instances are different to their nearest SEER prototypes (e.g. PSA). }
\end{subfigure}%
\\\\
\begin{subfigure}{0.5\textwidth}
\centering
\includegraphics[width=0.75\textwidth]{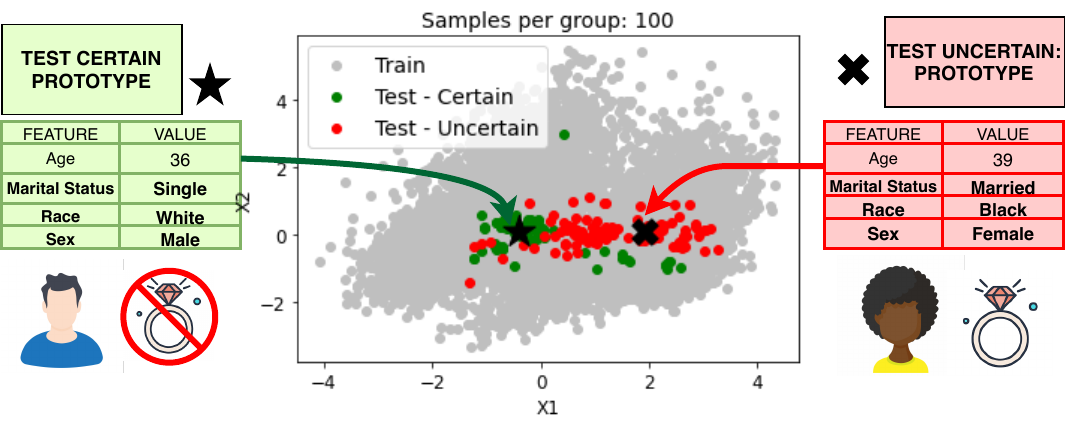}
\caption{ii. Adult: The \emph{certain} and \emph{uncertain} instances, represent two different demographics, aligning with the known dataset biases toward females. The \emph{uncertain} instances specifically highlight a sub-group of black, females.}
\end{subfigure}%
\\\\
\begin{subfigure}{0.5\textwidth}
\centering
\includegraphics[width=0.75\textwidth]{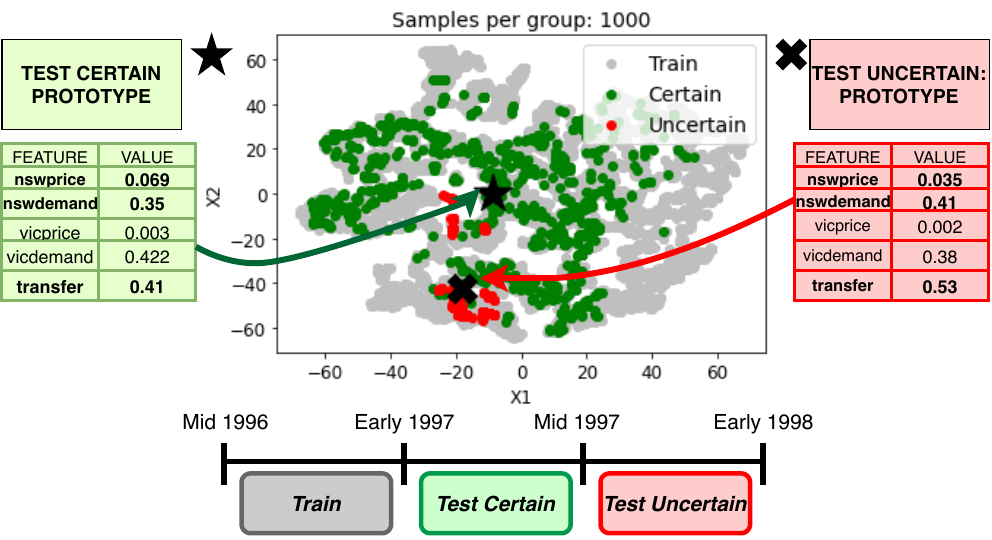}
\caption{iii. Electricity: The \emph{certain} instances are similar to the training set in features and time. The \emph{uncertain} instances identified represent a later time period, wherein concept drift has likely occurred.}
\end{subfigure}%
\caption{Insights of prototypes identified by Data-SUITE. Tables describe the average prototypes for \emph{certain} and \emph{uncertain} instances.}
\label{fig:insights}
\vspace{-0.5cm}
\end{figure}

For visual purposes, we embed the identified \emph{certain} and \emph{uncertain} instances into 2-D low-dimensional space as shown in Fig. \ref{fig:insights}. We clearly see that the \emph{certain} and \emph{uncertain} instances are distinct regions and that they lie in-distribution as evidenced by the embedding projection. This reinforces the quantitative findings of the previous experiment (i.e., that the identified instances are not OOD). We further highlight centroid,  ``average prototypes'' of the \emph{certain} and \emph{uncertain} regions as a digestible example of the region, which can easily be understood by stakeholders. For the SEER-CUTRACT analysis, in addition to prototypes for $\Dtest^{Cut}$ regions, we can also find the nearest neighbor SEER (USA) prototypes for each instance. Comparing the average and nearest neighbor prototypes assists us to tease out the incongruence between the two geographic sites.

Overall, Fig.~\ref{fig:insights}, in their respective captions, highlights the most valuable insights, quantitatively garnered on the basis of Data-SUITE across all three datasets. We conduct a more detailed analysis of the regions in Appendix \ref{app:eda}, to outline further potential practitioner usage.

\section{Discussion} \label{sec:discussion}
Automation should not replace the expertise and judgment of a data scientist in understanding the data, nor will it replace the ingenuity required to build better models. In this spirit, we developed Data-SUITE and illustrated its capability, across multiple datasets, to empower data scientists to perform more insightful data exploration, as well as, enable more reliable model deployment. 
We address these use-cases for the understudied problem of in-distribution heterogeneity and propose a flexible data-centric solution. This permits the identification of impactful instances, independent of a task-specific model. 
Data-SUITE allows to perform stratification of test data into inconsistent and uncertain instances with respect to training data.
This stratification has been shown to be in line with downstream performance and to provide valuable insights for profiling incongruent test instances in a rigorous and quantitative way. 
The quality of this stratification by Data-SUITE is not matched by any benchmark uncertainty estimation method (data-centric or not). 
The promising result serves two roles. 1. The method could be used by practitioners both to improve data exploration, as well as, enable more reliable model deployment. 2. Data-SUITE
opens up future avenues to advance the data-centric AI research agenda, taking it a step further by both explaining why instances might be classed as uncertain or inconsistent. Further, how this information could be leveraged either to correct the identified instances or improve the data collection process to improve overall data quality.
Finally, the current formulation has focused on tabular data. For usage with high-dimensional data, we refer the reader to Appendix \ref{appendix:expand} for proposals on possible modifications to Data-SUITE.

\section*{Acknowledgements}
The authors are grateful to Fergus Imrie, Zhaozhi Qian, Yuchao Qin, Krzysztof Kacprzyk, Kamile Stankeviciute and the 4 anonymous ICML reviewers for their useful comments \& feedback. Nabeel Seedat would like to acknowledge Hameeda Saif for her constant support and feedback. Nabeel Seedat is funded by the Cystic Fibrosis Trust, Jonathan Crabbé by Aviva and Mihaela van der Schaar by the Office of Naval Research (ONR), NSF 1722516.

\clearpage
\bibliographystyle{icml2022}
\bibliography{main}

\begin{thebibliography}{66}
\providecommand{\natexlab}[1]{#1}
\providecommand{\url}[1]{\texttt{#1}}
\expandafter\ifx\csname urlstyle\endcsname\relax
  \providecommand{\doi}[1]{doi: #1}\else
  \providecommand{\doi}{doi: \begingroup \urlstyle{rm}\Url}\fi

\bibitem[Alaa \& Van Der~Schaar(2020)Alaa and Van
  Der~Schaar]{alaa2020discriminative}
Alaa, A. and Van Der~Schaar, M.
\newblock Discriminative jackknife: Quantifying uncertainty in deep learning
  via higher-order influence functions.
\newblock In \emph{International Conference on Machine Learning}, pp.\
  165--174. PMLR, 2020.

\bibitem[Algan \& Ulusoy(2021)Algan and Ulusoy]{algan2021image}
Algan, G. and Ulusoy, I.
\newblock Image classification with deep learning in the presence of noisy
  labels: A survey.
\newblock \emph{Knowledge-Based Systems}, 215:\penalty0 106771, 2021.

\bibitem[Asuncion \& Newman(2007)Asuncion and Newman]{asuncion2007uci}
Asuncion, A. and Newman, D.
\newblock Uci machine learning repository, 2007.

\bibitem[Balasubramanian et~al.(2014)Balasubramanian, Ho, and
  Vovk]{balasubramanian2014conformal}
Balasubramanian, V., Ho, S.-S., and Vovk, V.
\newblock \emph{Conformal prediction for reliable machine learning: theory,
  adaptations and applications}.
\newblock Newnes, 2014.

\bibitem[Bedford \& Cooke(2001)Bedford and Cooke]{bedford2001probability}
Bedford, T. and Cooke, R.~M.
\newblock Probability density decomposition for conditionally dependent random
  variables modeled by vines.
\newblock \emph{Annals of Mathematics and Artificial intelligence}, 32\penalty0
  (1):\penalty0 245--268, 2001.

\bibitem[Borisov et~al.(2021)Borisov, Leemann, Se{\ss}ler, Haug, Pawelczyk, and
  Kasneci]{borisov2021deep}
Borisov, V., Leemann, T., Se{\ss}ler, K., Haug, J., Pawelczyk, M., and Kasneci,
  G.
\newblock Deep neural networks and tabular data: A survey.
\newblock \emph{arXiv preprint arXiv:2110.01889}, 2021.

\bibitem[Bostr{\"o}m et~al.(2016)Bostr{\"o}m, Linusson, L{\"o}fstr{\"o}m, and
  Johansson]{bostrom2016evaluation}
Bostr{\"o}m, H., Linusson, H., L{\"o}fstr{\"o}m, T., and Johansson, U.
\newblock Evaluation of a variance-based nonconformity measure for regression
  forests.
\newblock In \emph{Symposium on Conformal and Probabilistic Prediction with
  Applications}, pp.\  75--89. Springer, 2016.

\bibitem[Breunig et~al.(2000)Breunig, Kriegel, Ng, and Sander]{breunig2000lof}
Breunig, M.~M., Kriegel, H.-P., Ng, R.~T., and Sander, J.
\newblock Lof: identifying density-based local outliers.
\newblock In \emph{Proceedings of the 2000 ACM SIGMOD international conference
  on Management of data}, pp.\  93--104, 2000.

\bibitem[Chan et~al.(2020)Chan, Alaa, Qian, and Van
  Der~Schaar]{chan2020unlabelled}
Chan, A., Alaa, A., Qian, Z., and Van Der~Schaar, M.
\newblock Unlabelled data improves bayesian uncertainty calibration under
  covariate shift.
\newblock In \emph{International Conference on Machine Learning}, pp.\
  1392--1402. PMLR, 2020.

\bibitem[Duggan et~al.(2016)Duggan, Anderson, Altekruse, Penberthy, and
  Sherman]{duggan2016surveillance}
Duggan, M.~A., Anderson, W.~F., Altekruse, S., Penberthy, L., and Sherman,
  M.~E.
\newblock The surveillance, epidemiology and end results (seer) program and
  pathology: towards strengthening the critical relationship.
\newblock \emph{The American journal of surgical pathology}, 40\penalty0
  (12):\penalty0 e94, 2016.

\bibitem[Gal \& Ghahramani(2016)Gal and Ghahramani]{gal2016dropout}
Gal, Y. and Ghahramani, Z.
\newblock Dropout as a bayesian approximation: Representing model uncertainty
  in deep learning.
\newblock In \emph{international conference on machine learning}, pp.\
  1050--1059. PMLR, 2016.

\bibitem[Ghosh et~al.(2018)Ghosh, Yao, and Doshi-Velez]{ghosh2018structured}
Ghosh, S., Yao, J., and Doshi-Velez, F.
\newblock Structured variational learning of bayesian neural networks with
  horseshoe priors.
\newblock In \emph{International Conference on Machine Learning}, pp.\
  1744--1753. PMLR, 2018.

\bibitem[Gianfrancesco et~al.(2018)Gianfrancesco, Tamang, Yazdany, and
  Schmajuk]{gianfrancesco2018potential}
Gianfrancesco, M.~A., Tamang, S., Yazdany, J., and Schmajuk, G.
\newblock Potential biases in machine learning algorithms using electronic
  health record data.
\newblock \emph{JAMA internal medicine}, 178\penalty0 (11):\penalty0
  1544--1547, 2018.

\bibitem[Goodfellow et~al.(2014{\natexlab{a}})Goodfellow, Pouget-Abadie, Mirza,
  Xu, Warde-Farley, Ozair, Courville, and Bengio]{goodfellow}
Goodfellow, I., Pouget-Abadie, J., Mirza, M., Xu, B., Warde-Farley, D., Ozair,
  S., Courville, A., and Bengio, Y.
\newblock Generative adversarial nets.
\newblock In \emph{Advances in Neural Information Processing Systems 2014},
  volume~27, 2014{\natexlab{a}}.

\bibitem[Goodfellow et~al.(2014{\natexlab{b}})Goodfellow, Shlens, and
  Szegedy]{goodfellow2014explaining}
Goodfellow, I.~J., Shlens, J., and Szegedy, C.
\newblock Explaining and harnessing adversarial examples.
\newblock \emph{arXiv preprint arXiv:1412.6572}, 2014{\natexlab{b}}.

\bibitem[Graves(2011)]{graves2011practical}
Graves, A.
\newblock Practical variational inference for neural networks.
\newblock \emph{Advances in neural information processing systems}, 24, 2011.

\bibitem[Gretton et~al.(2012)Gretton, Borgwardt, Rasch, Sch{\"o}lkopf, and
  Smola]{gretton2012kernel}
Gretton, A., Borgwardt, K.~M., Rasch, M.~J., Sch{\"o}lkopf, B., and Smola, A.
\newblock A kernel two-sample test.
\newblock \emph{The Journal of Machine Learning Research}, 13\penalty0
  (1):\penalty0 723--773, 2012.

\bibitem[Gulrajani et~al.(2017)Gulrajani, Ahmed, Arjovsky, Dumoulin, and
  Courville]{gulrajani2017improved}
Gulrajani, I., Ahmed, F., Arjovsky, M., Dumoulin, V., and Courville, A.
\newblock Improved training of wasserstein gans.
\newblock In \emph{Proceedings of the 31st International Conference on Neural
  Information Processing Systems}, pp.\  5769--5779, 2017.

\bibitem[Gurumoorthy et~al.(2019)Gurumoorthy, Dhurandhar, Cecchi, and
  Aggarwal]{gurumoorthy2019efficient}
Gurumoorthy, K.~S., Dhurandhar, A., Cecchi, G., and Aggarwal, C.
\newblock Efficient data representation by selecting prototypes with importance
  weights.
\newblock In \emph{2019 IEEE International Conference on Data Mining (ICDM)},
  pp.\  260--269. IEEE, 2019.

\bibitem[Haeusser et~al.(2017)Haeusser, Frerix, Mordvintsev, and
  Cremers]{haeusser2017associative}
Haeusser, P., Frerix, T., Mordvintsev, A., and Cremers, D.
\newblock Associative domain adaptation.
\newblock In \emph{Proceedings of the IEEE international conference on computer
  vision}, pp.\  2765--2773, 2017.

\bibitem[Harries \& Wales(1999)Harries and Wales]{harries1999splice}
Harries, M. and Wales, N.~S.
\newblock Splice-2 comparative evaluation: Electricity pricing.
\newblock 1999.

\bibitem[Hsu et~al.(2020)Hsu, Shen, Jin, and Kira]{hsu2020generalized}
Hsu, Y.-C., Shen, Y., Jin, H., and Kira, Z.
\newblock Generalized odin: Detecting out-of-distribution image without
  learning from out-of-distribution data.
\newblock In \emph{Proceedings of the IEEE/CVF Conference on Computer Vision
  and Pattern Recognition}, pp.\  10951--10960, 2020.

\bibitem[Jain et~al.(2020)Jain, Patel, Nagalapatti, Gupta, Mehta, Guttula,
  Mujumdar, Afzal, Sharma~Mittal, and Munigala]{jain2020overview}
Jain, A., Patel, H., Nagalapatti, L., Gupta, N., Mehta, S., Guttula, S.,
  Mujumdar, S., Afzal, S., Sharma~Mittal, R., and Munigala, V.
\newblock Overview and importance of data quality for machine learning tasks.
\newblock In \emph{Proceedings of the 26th ACM SIGKDD International Conference
  on Knowledge Discovery \& Data Mining}, pp.\  3561--3562, 2020.

\bibitem[Joe(2014)]{joe2014dependence}
Joe, H.
\newblock \emph{Dependence modeling with copulas}.
\newblock CRC press, 2014.

\bibitem[Johansson et~al.(2015)Johansson, Sönströd, and Linusson]{Johansson}
Johansson, U., Sönströd, C., and Linusson, H.
\newblock Efficient conformal regressors using bagged neural nets.
\newblock In \emph{2015 International Joint Conference on Neural Networks
  (IJCNN)}, pp.\  1--8, 2015.
\newblock \doi{10.1109/IJCNN.2015.7280763}.

\bibitem[Kandel et~al.(2012)Kandel, Paepcke, Hellerstein, and
  Heer]{kandel2012enterprise}
Kandel, S., Paepcke, A., Hellerstein, J.~M., and Heer, J.
\newblock Enterprise data analysis and visualization: An interview study.
\newblock \emph{IEEE Transactions on Visualization and Computer Graphics},
  18\penalty0 (12):\penalty0 2917--2926, 2012.

\bibitem[Kingma \& Welling(2014)Kingma and Welling]{KingmaW13}
Kingma, D.~P. and Welling, M.
\newblock Auto-encoding variational bayes.
\newblock In \emph{2nd International Conference on Learning Representations},
  2014.

\bibitem[Kingma et~al.(2015)Kingma, Salimans, and
  Welling]{kingma2015variational}
Kingma, D.~P., Salimans, T., and Welling, M.
\newblock Variational dropout and the local reparameterization trick.
\newblock \emph{Advances in neural information processing systems},
  28:\penalty0 2575--2583, 2015.

\bibitem[Koenker \& Hallock(2001)Koenker and Hallock]{koenker2001quantile}
Koenker, R. and Hallock, K.~F.
\newblock Quantile regression.
\newblock \emph{Journal of economic perspectives}, 15\penalty0 (4):\penalty0
  143--156, 2001.

\bibitem[Kumagai \& Iwata(2019)Kumagai and Iwata]{kumagai2019unsupervised}
Kumagai, A. and Iwata, T.
\newblock Unsupervised domain adaptation by matching distributions based on the
  maximum mean discrepancy via unilateral transformations.
\newblock In \emph{Proceedings of the AAAI Conference on Artificial
  Intelligence}, volume~33, pp.\  4106--4113, 2019.

\bibitem[Kumar et~al.(2017)Kumar, Boehm, and Yang]{kumar2017data}
Kumar, A., Boehm, M., and Yang, J.
\newblock Data management in machine learning: Challenges, techniques, and
  systems.
\newblock In \emph{Proceedings of the 2017 ACM International Conference on
  Management of Data}, pp.\  1717--1722, 2017.

\bibitem[Lakshminarayanan et~al.(2017)Lakshminarayanan, Pritzel, and
  Blundell]{lakshminarayanan2017simple}
Lakshminarayanan, B., Pritzel, A., and Blundell, C.
\newblock Simple and scalable predictive uncertainty estimation using deep
  ensembles.
\newblock \emph{Advances in neural information processing systems}, 30, 2017.

\bibitem[Lecu{\'e} et~al.(2020)Lecu{\'e}, Lerasle, and
  Mathieu]{lecue2020robust}
Lecu{\'e}, G., Lerasle, M., and Mathieu, T.
\newblock Robust classification via mom minimization.
\newblock \emph{Machine Learning}, 109\penalty0 (8):\penalty0 1635--1665, 2020.

\bibitem[Lee et~al.(2018)Lee, Lee, Lee, and Shin]{lee2018simple}
Lee, K., Lee, K., Lee, H., and Shin, J.
\newblock A simple unified framework for detecting out-of-distribution samples
  and adversarial attacks.
\newblock \emph{Advances in neural information processing systems}, 31, 2018.

\bibitem[Lei et~al.(2018)Lei, G’Sell, Rinaldo, Tibshirani, and
  Wasserman]{lei2018distribution}
Lei, J., G’Sell, M., Rinaldo, A., Tibshirani, R.~J., and Wasserman, L.
\newblock Distribution-free predictive inference for regression.
\newblock \emph{Journal of the American Statistical Association}, 113\penalty0
  (523):\penalty0 1094--1111, 2018.

\bibitem[Leslie et~al.(2021)Leslie, Mazumder, Peppin, Wolters, and
  Hagerty]{leslie2021does}
Leslie, D., Mazumder, A., Peppin, A., Wolters, M.~K., and Hagerty, A.
\newblock Does “ai” stand for augmenting inequality in the era of covid-19
  healthcare?
\newblock \emph{BMJ}, 372, 2021.

\bibitem[Li et~al.(2020)Li, Zhao, Botta, Ionescu, and Hu]{li2020copod}
Li, Z., Zhao, Y., Botta, N., Ionescu, C., and Hu, X.
\newblock Copod: copula-based outlier detection.
\newblock In \emph{2020 IEEE International Conference on Data Mining (ICDM)},
  pp.\  1118--1123. IEEE, 2020.

\bibitem[Liu et~al.(2012)Liu, Ting, and Zhou]{liu2012isolation}
Liu, F.~T., Ting, K.~M., and Zhou, Z.-H.
\newblock Isolation-based anomaly detection.
\newblock \emph{ACM Transactions on Knowledge Discovery from Data (TKDD)},
  6\penalty0 (1):\penalty0 1--39, 2012.

\bibitem[Long et~al.(2015)Long, Cao, Wang, and Jordan]{long2015learning}
Long, M., Cao, Y., Wang, J., and Jordan, M.
\newblock Learning transferable features with deep adaptation networks.
\newblock In \emph{International conference on machine learning}, pp.\
  97--105. PMLR, 2015.

\bibitem[Navratil et~al.(2020)Navratil, Arnold, and
  Elder]{navratil2020uncertainty}
Navratil, J., Arnold, M., and Elder, B.
\newblock Uncertainty prediction for deep sequential regression using meta
  models.
\newblock \emph{arXiv preprint arXiv:2007.01350}, 2020.

\bibitem[Ng et~al.(2021)Ng, Aroyo, Coleman, Diamos, Reddi, Vanschoren, Wu, and
  Zhou]{ng2021}
Ng, A., Aroyo, L., Coleman, C., Diamos, G., Reddi, V.~J., Vanschoren, J., Wu,
  C.-J., and Zhou, S.
\newblock Neurips data-centric ai workshop, 2021.
\newblock URL \url{https://datacentricai.org/}.

\bibitem[Obermeyer et~al.(2019)Obermeyer, Powers, Vogeli, and
  Mullainathan]{obermeyer2019dissecting}
Obermeyer, Z., Powers, B., Vogeli, C., and Mullainathan, S.
\newblock Dissecting racial bias in an algorithm used to manage the health of
  populations.
\newblock \emph{Science}, 366\penalty0 (6464):\penalty0 447--453, 2019.

\bibitem[Park et~al.(2021)Park, Awadalla, Kohno, and Patel]{park2021reliable}
Park, C., Awadalla, A., Kohno, T., and Patel, S.
\newblock Reliable and trustworthy machine learning for health using dataset
  shift detection.
\newblock \emph{Advances in Neural Information Processing Systems}, 34, 2021.

\bibitem[Polyzotis et~al.(2017)Polyzotis, Roy, Whang, and
  Zinkevich]{polyzotis2017data}
Polyzotis, N., Roy, S., Whang, S.~E., and Zinkevich, M.
\newblock Data management challenges in production machine learning.
\newblock In \emph{Proceedings of the 2017 ACM International Conference on
  Management of Data}, pp.\  1723--1726, 2017.

\bibitem[Prostate Cancer~UK()]{prostate}
Prostate Cancer~UK, C.
\newblock Prostate cancer uk.
\newblock URL \url{https://prostatecanceruk.org/}.

\bibitem[Ren et~al.(2018)Ren, Zeng, Yang, and Urtasun]{ren2018learning}
Ren, M., Zeng, W., Yang, B., and Urtasun, R.
\newblock Learning to reweight examples for robust deep learning.
\newblock In \emph{International Conference on Machine Learning}, pp.\
  4334--4343. PMLR, 2018.

\bibitem[Rezende \& Mohamed(2015)Rezende and Mohamed]{rezende2015variational}
Rezende, D. and Mohamed, S.
\newblock Variational inference with normalizing flows.
\newblock In \emph{International conference on machine learning}, pp.\
  1530--1538. PMLR, 2015.

\bibitem[Sambasivan et~al.(2021)Sambasivan, Kapania, Highfill, Akrong,
  Paritosh, and Aroyo]{Sambasivan}
Sambasivan, N., Kapania, S., Highfill, H., Akrong, D., Paritosh, P.~K., and
  Aroyo, L.~M.
\newblock "everyone wants to do the model work, not the data work": Data
  cascades in high-stakes ai.
\newblock 2021.

\bibitem[Saria \& Subbaswamy(2019)Saria and Subbaswamy]{saria2019tutorial}
Saria, S. and Subbaswamy, A.
\newblock Tutorial: safe and reliable machine learning.
\newblock \emph{ACM Conference on Fairness, Accountability, and Transparency},
  2019.

\bibitem[Seedat \& Kanan(2019)Seedat and Kanan]{seedat2019towards}
Seedat, N. and Kanan, C.
\newblock Towards calibrated and scalable uncertainty representations for
  neural networks.
\newblock \emph{arXiv preprint arXiv:1911.00104}, 2019.

\bibitem[Shafer \& Vovk(2008)Shafer and Vovk]{shafer2008tutorial}
Shafer, G. and Vovk, V.
\newblock A tutorial on conformal prediction.
\newblock \emph{Journal of Machine Learning Research}, 9\penalty0 (3), 2008.

\bibitem[Sklar(1959)]{Skla59}
Sklar, A.
\newblock Fonctions de r\'epartition \`a n dimensions et leurs marges.
\newblock \emph{Publications de l'Institut de Statistique de l'Universit\'e de
  Paris}, 8:\penalty0 229--231, 1959.

\bibitem[Song et~al.(2020)Song, Kim, Park, Shin, and Lee]{song2020learning}
Song, H., Kim, M., Park, D., Shin, Y., and Lee, J.-G.
\newblock Learning from noisy labels with deep neural networks: A survey.
\newblock \emph{arXiv preprint arXiv:2007.08199}, 2020.

\bibitem[Srivastava et~al.(2017)Srivastava, Valkov, Russell, Gutmann, and
  Sutton]{srivastava2017veegan}
Srivastava, A., Valkov, L., Russell, C., Gutmann, M.~U., and Sutton, C.
\newblock Veegan: Reducing mode collapse in gans using implicit variational
  learning.
\newblock In \emph{Proceedings of the 31st International Conference on Neural
  Information Processing Systems}, pp.\  3310--3320, 2017.

\bibitem[Suciu et~al.(2011)Suciu, Olteanu, R{\'e}, and
  Koch]{suciu2011probabilistic}
Suciu, D., Olteanu, D., R{\'e}, C., and Koch, C.
\newblock Probabilistic databases.
\newblock \emph{Synthesis lectures on data management}, 3\penalty0
  (2):\penalty0 1--180, 2011.

\bibitem[Varshney(2020)]{varshney2020mismatched}
Varshney, K.~R.
\newblock On mismatched detection and safe, trustworthy machine learning.
\newblock In \emph{2020 54th Annual Conference on Information Sciences and
  Systems (CISS)}, pp.\  1--4. IEEE, 2020.

\bibitem[Vovk(2013)]{vovk2013transductive}
Vovk, V.
\newblock Transductive conformal predictors.
\newblock In \emph{IFIP International Conference on Artificial Intelligence
  Applications and Innovations}, pp.\  348--360. Springer, 2013.

\bibitem[Vovk et~al.(2005)Vovk, Gammerman, and Shafer]{vovk2005conformal}
Vovk, V., Gammerman, A., and Shafer, G.
\newblock Conformal prediction.
\newblock \emph{Algorithmic learning in a random world}, pp.\  17--51, 2005.

\bibitem[Wasserman(2004)]{wasserman2004all}
Wasserman, L.
\newblock \emph{All of statistics: a concise course in statistical inference},
  volume~26.
\newblock Springer, 2004.

\bibitem[Williams \& Rasmussen(2006)Williams and
  Rasmussen]{williams2006gaussian}
Williams, C.~K. and Rasmussen, C.~E.
\newblock \emph{Gaussian processes for machine learning}, volume~2.
\newblock MIT press Cambridge, MA, 2006.

\bibitem[Yan et~al.(2017)Yan, Ding, Li, Wang, Xu, and Zuo]{yanmmd}
Yan, H., Ding, Y., Li, P., Wang, Q., Xu, Y., and Zuo, W.
\newblock Mind the class weight bias: Weighted maximum mean discrepancy for
  unsupervised domain adaptation.
\newblock In \emph{2017 IEEE Conference on Computer Vision and Pattern
  Recognition (CVPR)}, pp.\  945--954, 2017.
\newblock \doi{10.1109/CVPR.2017.107}.

\bibitem[Yang et~al.(2021)Yang, Zhou, Li, and Liu]{yang2021generalized}
Yang, J., Zhou, K., Li, Y., and Liu, Z.
\newblock Generalized out-of-distribution detection: A survey.
\newblock \emph{arXiv preprint arXiv:2110.11334}, 2021.

\bibitem[Yoon et~al.(2020)Yoon, Zhang, Jordon, and van~der
  Schaar]{yoon2020vime}
Yoon, J., Zhang, Y., Jordon, J., and van~der Schaar, M.
\newblock Vime: Extending the success of self-and semi-supervised learning to
  tabular domain.
\newblock \emph{Advances in Neural Information Processing Systems}, 33, 2020.

\bibitem[Zhang et~al.(2021)Zhang, Goldstein, and
  Ranganath]{zhang2021understanding}
Zhang, L., Goldstein, M., and Ranganath, R.
\newblock Understanding failures in out-of-distribution detection with deep
  generative models.
\newblock In \emph{International Conference on Machine Learning}, pp.\
  12427--12436. PMLR, 2021.

\bibitem[Zhao et~al.(2021)Zhao, Hu, Cheng, Wang, Wan, Wang, Yang, Bai, Li,
  Xiao, et~al.]{zhao2021suod}
Zhao, Y., Hu, X., Cheng, C., Wang, C., Wan, C., Wang, W., Yang, J., Bai, H.,
  Li, Z., Xiao, C., et~al.
\newblock Suod: Accelerating large-scale unsupervised heterogeneous outlier
  detection.
\newblock \emph{Proceedings of Machine Learning and Systems}, 3, 2021.

\bibitem[Zliobaite(2013)]{zliobaite2013good}
Zliobaite, I.
\newblock How good is the electricity benchmark for evaluating concept drift
  adaptation.
\newblock \emph{arXiv preprint arXiv:1301.3524}, 2013.

\end{thebibliography}

\clearpage
\onecolumn
\appendix
\section{Data-SUITE details \& related work} \label{sec:appendixA}
\subsection{Extended Related Work}

We present a comparison of our framework Data-SUITE, and contrast it to the related work of uncertainty quantification and learning with noisy labels. Table \ref{related_work}, highlights both similarities and differences across multiple dimensions. We highlight 3 key features which distinguish Data-SUITE: (1) Data-centric uncertainty is a novel paradigm compared to the predominant model-centric approaches, (2) Our method offers increased flexibility, as it is used independent of task-specific predictive models. Any conclusions that we draw from Data-SUITE are not model-specific. (3) Our method provides theoretical guarantees concerning the validity of coverage.

\begin{table*}[!h]
\centering
\caption{Comparison of related work}
\vspace{0.1in}
\scalebox{1}{
\begin{tabular}{@{}lcccccc@{}}
\toprule
      & \makecell{Data-centric \\ uncertainty}    & \makecell{Model-centric \\ uncertainty}  &  \makecell{Task Model \\ independent}  & \makecell{No noise \\ assumptions} & \makecell{Coverage \\guarantees}                          \\ \midrule
Data-SUITE (Ours) & $\checkmark$ & $\times$ & $\checkmark$    & $\checkmark$  & $\checkmark$ \\
Uncertainty Quantification    & $\times$   & $\checkmark$ & $\times$    & $\checkmark$ & $\times$ \\
Noisy labels &   $\times$ & $\checkmark$  & $\times$   & $\times$ & $\times$ \\ \bottomrule
\end{tabular}}
\label{related_work}
\end{table*}

\subsection{Data-SUITE Details}
\begin{figure}[!h]
    \centering
    \includegraphics[width=\linewidth]{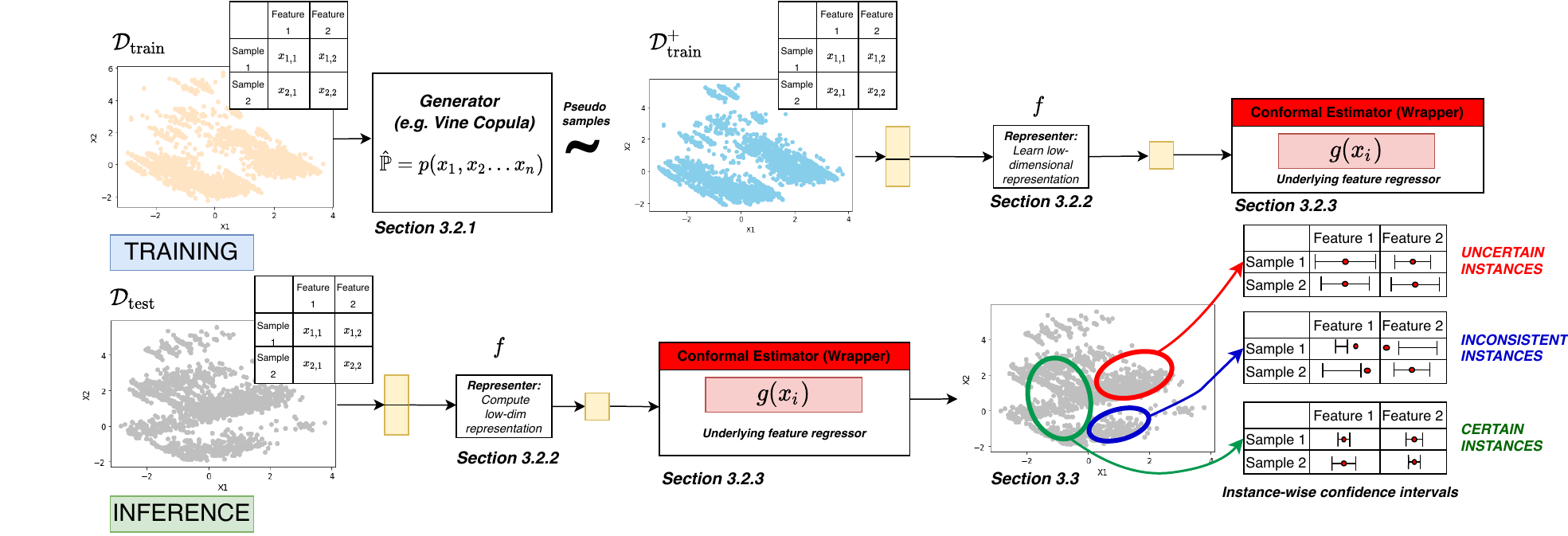}
    \caption{Outline of our framework \textbf{Data-SUITE}}
    \label{fig:flow_diagram}
\end{figure}

We present a block diagram of our framework Data-SUITE in Figure \ref{fig:flow_diagram}. We next have in-depth discussions on both the generator and conformal predictor. We outline the motivations as well as technical details not covered in the main paper.

\subsubsection{Generator: Copulas}\label{appendix:copula}

\paragraph{Motivation.}
Recall in our formulation, we have a set of training instances $\Dtrain$ and we wish to learn the dependency between features $\X$. Hence, we model the multivariate joint distribution of $\Dtrain$ to capture the dependence structure between random variables. A significant challenge is modeling distributions in high dimensions. Parametric approaches such as Kernel Density prediction (KDE), often using Gaussian distributions, are largely inflexible.

On the other hand, nonparametric versions are typically infeasible with complex data distributions and the curse of dimensionality.
Additionally, while Variational Autoencoders (VAEs) \cite{KingmaW13} and Generative Adversarial Networks (GANs) \cite{goodfellow} can model learn the joint distributions, they both have limitations for our setting. VAEs make strong distributional assumptions \cite{rezende2015variational}, while GANs involve training multiple models, which leads to associated difficulties \cite{srivastava2017veegan,gulrajani2017improved} (both from training computational burden and inherent to GANs - e.g., mode collapse). 

An attractive approach, particularly for tabular data, is Copulas; which flexibly couple the marginal distributions of the different random variables into a joint distribution. One important reason lies in the following theorem:
\begin{property}[Sklar’s theorem]
A d-dimensional random vector \textbf{X} = ($X_1,...,X_d$) with joint
distribution $F$ and marginal distributions $F_1,...,F_d$ can be expressed as $f(X_1,...,X_d)=C_{\theta}\{F_1(X_1)...F_d(X_d)\}$, where $C_{\theta}: [0,1]^{d} \rightarrow [0,1]$ is a Copula. This is attractive in high dimensions, as it separates the learning of univariate marginal distributions from learning of the coupled multivariate dependence structure.
\end{property}

Parametric copulas have limited expressivity; hence we use pair copula constructions (vine copulas) \cite{bedford2001probability}, which are hierarchical and express a multivariate copula as a cascade of bivariate copulas. 

\paragraph{Copula Details.}
To learn the copula, we factorize the d-dimensional copula density into a product of $d(d-1)/2$, bivariate conditional densities by means of the chain rule. 

The graphical model which has edges being each bivariate-copula that encodes the (conditional) dependence between variables. The graphical model additionally consists of levels (where there are as many levels as features of the dataset). Each node will be a variable, and edges are coupling between variables based on bivariate copulas. As each level is constructed, the number of nodes decreases per level. The product over all pair-copula densities is then taken to define the joint copula.

Once we have learned the copula density, we sample the copula to obtain an augmented dataset of pseudo/synthetic samples. The copula samples are then easily transformed back to the natural data scale using the inverse probability integral transform. 

Specifically, assume we have $\mathbf{U}= (U_1,U_2...U_d)$ random variable $~U(0,1)$. We can then use the Copula $C_{\theta}$ to define variables $\mathbf{S}= (S_1,S_2...S_d)$, where $S_1 = C^{-1}(U1), S_2 = C^{-1}(U2|U1)...S_d=C^{-1}(Ud|U1,U2...U_{d-1})$. This means that $\mathbf{S}$ is the inverse Rosenblatt transform of $\mathbf{U}$ and hence, $S \sim C$, which allows us to simulate synthetic/pseudo samples. For more on Copulas in general, we refer the reader to \cite{joe2014dependence}.

\paragraph{Complexity.} As we go through each tree, there are a decreasing number of pair copulas. i.e. ($T_1 = d-1$, $T_2=d-2...T_{d-1}=1$). Hence, the  complexity of this algorithm is $O(n) \times d \times \mbox{truncation level})$, which for all purposes is $O(n)$.

\subsubsection{Conformal predictor}\label{appendix:conformal}
\paragraph{Motivation.}
Conformal prediction allows us to transform any underlying point predictor into a valid interval predictor. We will not discuss the generalized framework of conformal prediction (\textit{Transductive Conformal prediction}), which requires model training to be redone for every data point. This has a large computational burden for modern datasets with many datapoints. For more, see \cite{shafer2008tutorial,balasubramanian2014conformal}.

We instead only discuss \emph{Inductive Conformal prediction}, which is used in Data-SUITE. The inductive method splits the two processes needed: 1. the training of the underlying model and 2. computing the conformal estimates. 

\paragraph{Conformal Prediction Details.}
Practically, we split the training set ($|\Dtrain^+|=n$) into two disjoint sets, namely the proper training set and calibration set: $\Dtrain^+ = \Dtrainb^+ \sqcup \Dcal^+$, where $|\Dtrainb^+|=m$ and ($|\Dcal^+|=n-m$).

We use the proper training set to create our prediction rules for the feature-wise regressor ($g$).The calibration set is used for ``conformalization'', i.e. for computing the non-conformity scores and p-values. 

The non-conformity score $\mu_i$ of each example is a function which computes the disagreement (i.e. non-conformance)
between the prediction and the true value. Note, we only compute non-conformity scores on the calibration set.

To obtain our intervals, we need to determine the critical value $\epsilon$ based on the non-conformity scores. We first sort them in in descending order. The critical value $\epsilon$ is then the $\lceil(|\Dcal^+| + 1)(1-\alpha)\rceil$-th smallest residual \cite{vovk2013transductive}. Consequently, for any confidence level (1-$\alpha$), we can use the critical value to find $p(x) > \alpha$, which corresponds to the maximum and minimum values with p-values larger than $\alpha$. As a consequence, we can obtain the maximum and minimum values of our predictive intervals.  

The whole process is detailed in Algorithm \ref{algorithm:ic}. Normalization to obtain adaptive instance-wise intervals can be done using a normalization function $sigma_i$, as described in the main paper.

\newpage
\paragraph{Remarks on theoretical guarantees.}\label{app:conf_guarantees}
A motivation for conformal prediction in high-stakes settings is the theoretical guarantees on CI coverage validity (See Property \ref{validity}).

i.e., at a confidence level ($1-\alpha$) of 95\% ($\alpha=0.05$), the true value will be within the CIs in at least 95\% of the cases.  The framework is non-parametric and only makes the exchangeability assumption (which we detail next) and the guarantees of validity of coverage hold for any choice of dataset, underlying model or nonconformity measure – which makes it a versatile option. 
\begin{property}[Validity] 
 Under the exchangeability assumption, the conformal predictor will return an interval, $\mathbb{P}(Y \in [l_i(x), r_i(x)]) \geq 1-\alpha$, i.e, error $ \leq \alpha$.
 \label{validity}
\end{property}
The validity of the CI holds if the data is independent and identically distributed (IID). In practice, the weaker assumption of exchangeability (see Assumption \ref{exchangeability}) also guarantees validity \cite{lei2018distribution}. This means that we are not required to impose any additional requirements for the validity of the CI, since the aforementioned assumptions on the underlying data are typically made for any ML model.

We also highlight that the validity is maintained even in the case of normalization, as long as the proper training set and calibration set are disjoint.

\begin{assumption}[Exchangeability] 
In a dataset of $n$ observations, the data points do not follow any particular order, i.e., all $n$ permutations are equiprobable. Exchangeability is weaker than IID observations; however, IID observations satisfy exchangeability.
\label{exchangeability}
\end{assumption}

    \SetKwFunction{proc}{$\Gamma^\alpha$}
    \SetKwProg{myproc}{procedure}{}{}
    \SetKwBlock{DummyBlock}{}{}
    \begin{algorithm}[t!]
        \SetKwInOut{Input}{Input}
        \SetKwInOut{Output}{Output}
        \underline{function CE} $(\alpha,\Dtrainb^+,\Dcal^+,\mu)$\;
        
        \Input{Significance $\alpha$, nonconformity measure $\mu$, proper training set, $\Dtrainb^+$ and calibration set, $\Dcal^+$}
        \Output{Interval estimate\newline}
        Train the underlying model $g$ on $\Dtrainb^+$\;
        
        Compute non-conformity scores on the calibration set $\Dcal^+$
        
        $P = \{\}$ ;
        \ForEach{$(\mathbf{x},x_i) \in \Dcal^+ $}
        {$
            \mu_i(x) =  |x_i - g_i \circ f(\mathbf{x})|$;
            
        Add $\mu_{i}(x)$ to P
        }
        
        \textbf{CONFORMALIZATION.}
        
        Sort P in descending order to obtain scores $S$
        
        Determine the critical value of $\epsilon \leftarrow
        \lceil(|\Dcal^+| + 1)(1-\alpha)\rceil$-th smallest residual in $S$. 
        
        Construct the interval interval predictor for each new value:\;
        \SetAlgoNoLine\DummyBlock{\SetAlgoLined
            \myproc{\proc{$\mathbf{x}:\mathcal{X}$}}{
            \ForEach{$ \mathbf{x} \in \Dtest$}{
                Apply $x_{m} = g_i \circ f(\mathbf{x}$)
            }
            \KwRet$[l_i(x), r_i(x)] = [x_{m}-\epsilon, x_{m}+\epsilon]$\;
            }
        }
        \caption{General Inductive Conformal prediction}
        \label{algorithm:ic}
    \end{algorithm}
    
\newpage
\subsubsection{Data-SUITE beyond tabular}\label{appendix:expand}

Currently, the focus of Data-SUITE has primarily engaged with tabular data being the most ubiquitous data type across industries \cite{borisov2021deep}. However, there is of course value in application to other high dimensional modalities such as images or text. We provide some possible ideas of how Data-SUITE could be adapted for these other modalities.

\paragraph{Generator.} As discussed in the main text, a copula might not be ideal for very high-dimensional (large $d_X$) data in domains such as computer vision or genomics. In those cases, copula modeling can be replaced by domain-specific augmentation techniques.  For example, Generative Adversarial Networks (GANs).

\paragraph{Representer.} PCA while the workhorse of tabular data, is typically not suitable for modalities such as images. Methods such as autoencoders could easily replace PCA in this instance.

\paragraph{Conformal Predictor.} The primary challenge lies in the conformal predictor. Data-SUITE builds feature-wise regressors – which fits the tabular feature-wise setting. Two alternatives are proposed for other modalities. 1. Instead of feature-wise regressors, to construct instance-wise regressors or 2. consider the components of an intermediate representation vector as the ``features''.
\clearpage
\section{Benchmarks \& Experimental Details} \label{sec:appendixB}
\subsection{Benchmarks \& Implementations}\label{appendix:benchmarks}

\subsubsection{Data-SUITE}
\paragraph{Implementation details.}
Data-SUITE adopts a pipeline-based approach to constructing CI's, leveraging copula modeling, representation learning and conformal estimation. 

We break down each of these below:\\
\textit{Copula Modeling}: We use the Copula to estimate a multivariate distribution with univariate marginal distributions. We make use vine copulas \cite{bedford2001probability}. for this task.

Specifically, we use Direct-Vines (D-Vines), which impose constraints on the edges such that we only learn vines of the structure given in Figure \ref{vine} below. 
When fitting each tree, we select the ``best'' base copula (Gaussian, Frank, Clayton or Gumbel) based on likelihood.

\begin{figure}[!h]
    \centering
     \vspace{-5mm}
    \includegraphics[width=0.25\linewidth]{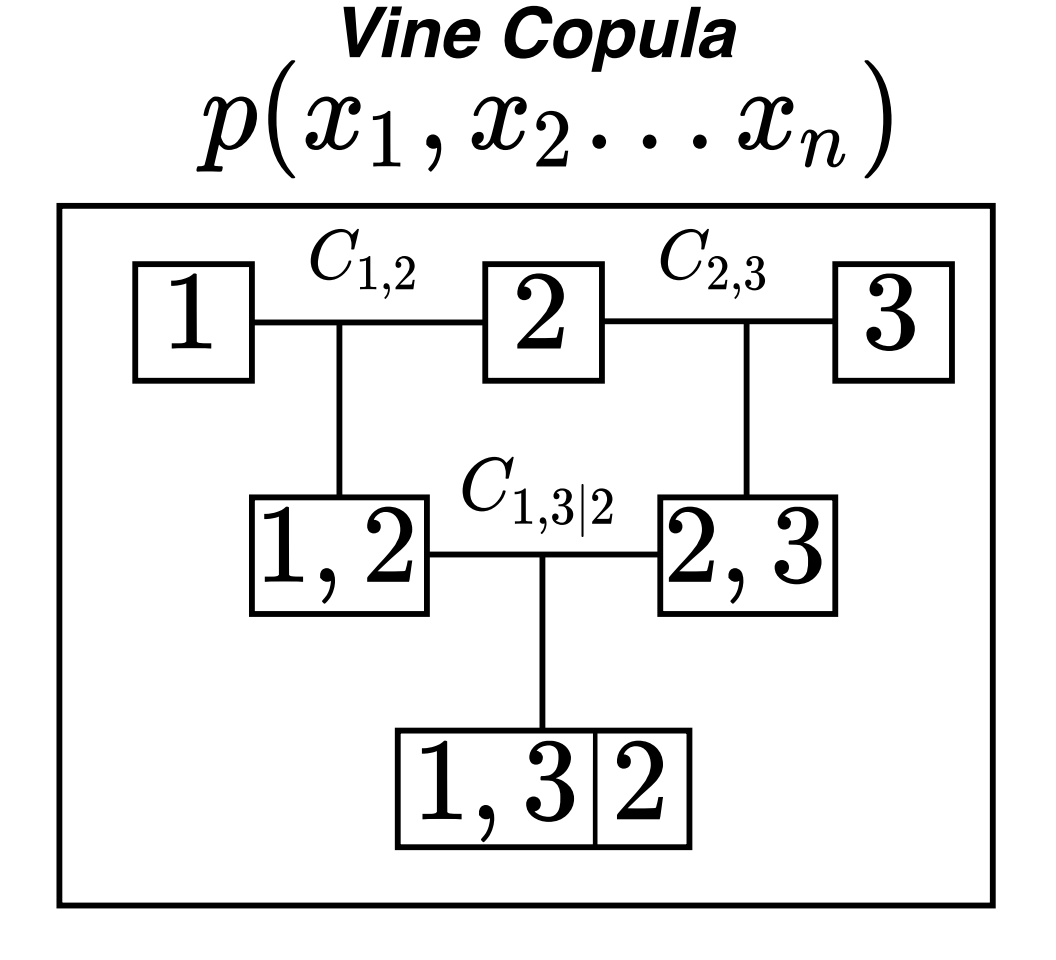}
    \vspace{-5mm}
    \caption{D-Vine structure}
    \label{vine}
\end{figure}

Finally, when we sample from the copula, we specifically sample $n_{samples} = |\Dtrain|$.

\textit{Representation learning}:
For the representer, we specifically make use of Principal Component Analysis (PCA). That said, we can simply replace this block with any alternative such as an autoencoder.

We pre-process all data prior to the representer, by standardizing the data, such that each feature has zero mean and unit variance (see Eq. \ref{standard}). Note if the features were indeed categorical, we perform one-hot encoding of the features.
\begin{align}
    z = \frac{(x - \mu)}{\sigma}
    \label{standard}
\end{align}

When applying PCA, to learn the latent representation, we halve the dimensionality of $\Dtrain$, i.e. $d_X/2$.

\textit{Conformal Prediction}:
The two most important design decisions for conformal estimation is selecting the underlying feature-wise regressor ($g_i$) and the non-conformity score ($\mu_i$).

We selected $g_i$, as in conventional machine learning by grid-search over different possible underlying models, evaluated on a validation set. We evaluate an MLP, KNN, Decision Trees and Random Forests. We ultimately selected a Decision Tree to serve as the base model for all $n=|d_X|$ feature-wise regessors. We use the following parameters max depth=None, min samples split=2, min samples leaf=5. 

Interestingly, a simpler base model proved to be more effective and outperformed more parameterized models, which would require more compute. This is an additional advantage of Data-SUITE. We motivate that the simpler model might be directly related to the fact that we build feature-wise regressors, hence the mapping function is easier to approximate.

Finally, we use the absolute error non-conformity score. In practice, we can use any non-conformity score, however, intuitively for our application where we want to approximate the true value as closely as possible, the absolute error makes the most sense. Our non-conformity score used is given by Eq. \ref{ncs} below.
\begin{align}
    \mu_i(x) =  |x_i - (g_i \circ f)(x)|
    \label{ncs}
\end{align}

\subsubsection{Bayesian Neural Network (BNN)}

Bayesian modelling, when applied to neural networks, involves a likelihood function $p(Y\,|\,X,\theta)$, where the parameters $\theta$ are estimated by a neural network. In contrast to conventional neural networks which have point estimates for the parameters, BNNs aim to learn distributions over parameters. However, modern neural networks have many parameters and weights. This makes exact inference largely intractable.

Hence, approximate Bayesian methods such as variational inference are often used instead. By this we mean we do not compute the posterior in the conventional manner. Rather, the variational approximation means we replace the posterior $p(\theta|\mathcal{D})$ with a more tractable variational distribution  $q(\theta;\lambda)$. The Kullback-Leibler (KL) divergence between the true distribution and the variational distribution is then minimized as the loss function, which equates to maximizing the evidence lower bound (ELBO), see Equation \ref{kl}. We note, however that the reparameterization trick \cite{kingma2015variational} is needed to make back propagation possible.
\begin{align}\label{kl}
    \mathcal{L}_{\text{VI}}\big(\mathcal{D};\lambda) := D_{\text{KL}}\big(q(\theta;\lambda)\,||\,p(\theta)\big) - \text{E}\big[\log p(\mathcal{D}\,|\,\theta)\big]
\end{align}

\paragraph{Implementation details.}  We train a 5-layer MLP model. A Gaussian prior is placed over the weights and we optimize the KL divergence during training. Our implementation is based on \cite{ghosh2018structured} and we use the implementation from \footnote{https://github.com/IBM/UQ360}.

\subsubsection{Deep Ensembles (ENSEMBLE)}

Deep Ensembles \cite{lakshminarayanan2017simple} is widely regarded as the state-of-the-art non-Bayesian uncertainty estimation method. The rationale is that training multiple randomly initialized models allows more robust predictions. Uncertainty can be computed as the variance of the different model predictions. We highlight some important features of Deep Ensembles: (1) optimization based on a  \textit{proper scoring rule} such that the loss has a unique minimum - which encourages the model to approximate the true probability distribution. However, the proper nature of the scoring rule introduces a distributional assumption. 
(2) Deep Ensembles uses adversarial perturbations based on the Fast Gradient Sign Method \cite{goodfellow2014explaining} and given by Equation  \ref{fgsm}.

\begin{align}
    \mathbf{x}' := \mathbf{x} + \boldsymbol{\eta}\odot\text{sgn}\big(\nabla_{\mathbf{x}}\mathcal{L}(\mathbf{x},y;\theta)\big)\label{fgsm}\,,
\end{align}
where $\odot$ denotes element-wise multiplication. 

This modifies the loss function used for gradient descent:
    \begin{align}
        \mathcal{L}_{\text{tot}}(\mathbf{X},\mathbf{y};\theta) := \mathcal{L}(\mathbf{X},\mathbf{y};\theta) + \mathcal{L}(\mathbf{X}',\mathbf{y};\theta)\,,
    \end{align}

To construct prediction intervals with an uncertainty estimate, an assumption of a conditionally normal distribution is assumed and the intervals, with uncertainty estimates are given as per Eq. \ref{de:uncert}.

\begin{align}
        \label{de:uncert}
        \Gamma(\mathbf{x}^*) := \left[\text{E}[y^*\,|\,\mathbf{x}^*] - \sqrt{var[y^*\,|\,\mathbf{x}^*]}, \text{E}[y^*\,|\,\mathbf{x}^*] + \sqrt{var[y^*\,|\,\mathbf{x}^*]}\right].
\end{align}

\paragraph{Implementation details.} We have 5 models in the ensemble, all randomly initialized. Each model is a 3-layer MLP, which we train for 10 epochs. The learning rate was empirically determined based on a validation set. To compute the uncertainty estimates at test time we obtain predictions from each model in the ensemble. The prediction interval is computed as per Eq. \ref{de:uncert}.

\subsubsection{Gaussian Process (GP)}
Gaussian process (GP) models ~\cite{williams1996gaussian} are fully  characterized by: the mean and covariance functions $\mathbf{\mu}$ and $\mathbf{\sum}$. The inference step can be performed exactly, as marginalizing the multivariate normal distributions can be written in closed form. See Equation \ref{gp}, whereby conditioning on the dataset $\mathcal{D}$, we can compute the posterior.

\begin{align}\label{gp}
    \left.\left(y^*\,\right|\,\mathbf{x}^*,\mathcal{D}\right)\sim\mathcal{N}\Big(\mu^* + (\Sigma^*)^t\Sigma^{-1}(\mathbf{y}-\boldsymbol{\mu}),\Sigma^{**} - (\Sigma^*)^t\Sigma^{-1}\Sigma^*\Big)\,.
\end{align}

We note the two assumptions made by GPs: (1) the data is conditionally normally distributed and (2) the covariance selection is correct. That said, violations to these assumptions can severely impact performance.  In addition, GPs have an issue of computational complexity due to matrix inversion of $O(n^3)$. For high-dimensional data, stochastic approximations are often used instead.

\paragraph{Implementation details.}  The GP is fit with a radial basis function (RBF) kernel and we make use of the Scikit-learn \footnote{https://scikit-learn.org} implementation of GPs which is based on \cite{williams2006gaussian}.   

\subsubsection{Monte-Carlo Dropout (MCD)}
Dropout layers are typically used as a regularizer at training time .~\cite{srivastava2014dropout}.  As shown by \cite{gal2016dropout} dropout networks can be used at test time (Monte-Carlo Dropout (MCD)) and are a variational approximation to deep Gaussian processes. Note that this induces the assumption of normality similar to GPs.

A key component of MCD, is that dropout at test time effectively allows us to obtain an ensemble of different models without having to retrain the model itself. For this, we do multiple stochastic forward passes at test time. 

When making predictions, we compute a conditional mean. This is approximated by Monte-Carlo integration, i.e. mean  of multiple forward passes. The prediction interval to characterize the uncertainty is given by Eq. \ref{mcd:uncert}:
\begin{align}
        \label{mcd:uncert}
        \Gamma(\mathbf{x}^*) := \left[\text{E}[y^*\,|\,\mathbf{x}^*] - \sqrt{var[y^*\,|\,\mathbf{x}^*]}, \text{E}[y^*\,|\,\mathbf{x}^*] + \sqrt{var[y^*\,|\,\mathbf{x}^*]}\right].
\end{align}

\paragraph{Implementation details.}  We train a 3-layer MLP with dropout ($p=0.1$)for 10 epochs. The learning rate was empirically determined based on a validation set. To compute uncertainty estimates at test time, we perform 20 Monte-Carlo Samples (forward passes at test time). The prediction interval is computed as per Eq. \ref{mcd:uncert}.

\subsubsection{Quantile Regression (QR)}
Quantile regressors estimate the conditional quantiles of a distribution (rather than conditional mean). Consequently, there is a natural measure of the underlying distribution's spread.

Typically, the MSE loss is replaced by the pinball loss, otherwise known as quantile loss, given by Equation \ref{pinball_loss}, which balances the number of points above and below the quantiles \cite{koenker2001quantile}. Neural networks are easily applied and optimize the loss function with the appropriate number of output heads.
    \begin{align}
        \label{pinball_loss}
        \mathcal{L}_{\text{pinball}}(\mathcal{T}) := \sum_{(\mathbf{x},y)\in\mathcal{T}} \max\Big((1-\alpha)(\hat{q}_\alpha(\mathbf{x})-y), \alpha(y-\hat{q}_\alpha(\mathbf{x}))\Big)\,.
    \end{align}

\paragraph{Implementation details.} The base model for predicting the quantiles is a Gradient Boosting Regressor, with 10 estimators, max depth=5,minimum samples per leaf=5, minimum samples for split=10.  We use the implementation from \footnote{https://github.com/IBM/UQ360}.

\newpage
\subsection{Synthetic Experiment Details}\label{appendix:synthetic}
\subsubsection{Dataset \& Experiment Configurations}

The synthetic data $\textbf{X} = [X_1, X_2, X_3]$ is drawn IID from a multivariate Gaussian distribution, parameterized by mean vector $\mu$ and a positive definite covariance matrix $\sum$. 

$\bm{\mu} = \begin{bmatrix}
5.0\\
0.0 \\
10.0
\end{bmatrix} \hspace{2cm}
\bm{\sum} = \begin{bmatrix}
3.40 & -2.75 & -2.00\\
-2.75 &  5.50 & 1.50 \\
-2.00 & 1.50 & 1.25
\end{bmatrix}$

We sample $n=1000$ for both $\Dtrain$ and $\Dtest$ respectively. We encode inconsistency and uncertainty into the features of the test set $\Dtest$ using a multivariate additive  model $\hat{X} = X+Z$.  where $Z \in R^{n \times m}$, is the perturbation matrix.  

We conduct three experiments with different configurations ($D_a,D_b ,D_c$), see Table \ref{tab:synthetic_paramterizations} 

\begin{table}[h]
    \centering
        \caption{Different configurations of the synthetic data}
    \scalebox{1}{
    \begin{tabular}{lccc}
\toprule
 &     Noise distribution & Perturbation Proportions & Perturbation variance   \\
\midrule
$D_a$ &  $\{Normal\}$   &  $\{0.1, 0.25, 0.5, 0.75\}$  & $\{2\}$  \\
$D_b$ &  $\{Normal\}$   &  $\{0.5\}$  & $\{1,2,3\}$  \\
$D_c$ &  $\{Beta, Gamma, Normal, Weibull\}$   &  $\{0.5\}$  & $\{2\}$  \\
\bottomrule
\end{tabular}}
    \label{tab:synthetic_paramterizations}
\end{table}

\subsubsection{Downstream Model}
In Section 4.2, we have a downstream task wherein we compute the MSE for different models. The base model which we train using $\Dtrain$ is a linear regression model.

\subsection{Real-Data Experiment Details}\label{appendix:real}

\subsubsection{Datasets}

\paragraph{SEER Dataset} 
The SEER dataset consists of 240,486 patients enrolled in the American SEER program~\cite{duggan2016surveillance}. The dataset consists of features used to characterize \textbf{prostate cancer}: including age, PSA (severity score), Gleason score, clinical stage, treatments etc. A summary of the covariate features can be found in Table \ref{tab:seer_features}. The classification task is to predict patient mortality, which is binary label $\in \{ 0 , 1 \}$.

The dataset is highly imbalanced, where $~94\%$ of patients survive. Hence, we extract a balanced subset of of 20,000 patients (i.e. 10,000 with label=0 and 10,000 with label=1).

\begin{table}[h]
\caption{Summary of features for the SEER Dataset \cite{duggan2016surveillance}}
\begin{tabular}{ll}
\toprule
Feature & Range \\ 
\midrule
Age & $37-95$ \\ 
PSA & $0-98$ \\ 
Comorbidities & $0, 1, 2, \geq 3$ \\ 
Treatment & Hormone Therapy (PHT), Radical Therapy - RDx (RT-RDx), \newline Radical Therapy -Sx (RT-Sx), CM \\ 
Grade & $1, 2, 3, 4, 5$ \\ 
Stage & $1, 2, 3, 4$ \\ 
Primary Gleason & $1, 2, 3, 4, 5$ \\ 
Secondary Gleason & $1, 2, 3, 4, 5$ \\ 
\bottomrule
\end{tabular} 
\label{tab:seer_features}
\end{table}

\paragraph{CUTRACT Dataset}

The CUTRACT dataset is a private dataset consisting of 10,086 patients enrolled in the British Prostate Cancer UK program~\cite{prostate}. Similar, to the SEER dataset, it consists of the same features to characterize prostate cancer. Additionally, it has the same task to predict mortality. A summary of the covariate features can be found in Table \ref{tab:cutract_features}.

Once again, the dataset is highly imbalanced, hence we then choose extract a balanced subset of of 2,000 patients (i.e. 1000 with label=0 and 1000 with label=1).

\begin{table}[h]
\caption{Summary of features for the CUTRACT Dataset \cite{prostate}}
\begin{tabular}{ll}
\toprule
Feature & Range \\ 
\midrule
Age & $44-95$ \\ 
PSA & $1-100$ \\ 
Comorbidities & $0, 1, 2, \geq 3$ \\ 
Treatment & Hormone Therapy (PHT), Radical Therapy - RDx (RT-RDx), \newline Radical Therapy -Sx (RT-Sx), CM \\ 
Grade & $1, 2, 3, 4, 5$ \\ 
Stage & $1, 2, 3, 4$ \\ 
Primary Gleason & $1, 2, 3, 4, 5$ \\ 
Secondary Gleason & $1, 2, 3, 4, 5$ \\ 
\bottomrule
\end{tabular} 
\vspace{.5cm}
\label{tab:cutract_features}
\end{table}

\paragraph{ADULT Dataset}
The ADULT dataset \cite{asuncion2007uci} has 32,561 instances with a total of 13 attributes capturing demographic (age, gender, race), personal (marital status) and financial (income) features amongst others. The classification task predicts whether a person earns over \$50K or not. We encode the features (e.g. race, sex, gender etc) and a summary can be found in Table \ref{tab:adult_features}.

There is a known bias between gender and income in the dataset. We perform a train-test split such that $\Dtrain$ and $\Dtest$ have approximately equal sizes, with 15,378 and 14,784 samples respectively. In particular, to highlight the data exploration use-case. 

Note that there is an imbalance across certain features. However, these are amongst the sensitive attributes. Thus, we do not want to balance the datasets based on this, as we wish to show both in the data exploration and model deployment experiments that we can identify these biases in the datasets. Balancing might eliminate these biases.

\begin{table}[h]
\centering
\caption{Summary of features for the ADULT Dataset \cite{asuncion2007uci}}
\begin{tabular}{ll}
\toprule
Feature & Range \\ 
\midrule
Age & $17-90$ \\ 
education-num & $1-16$ \\ 
marital-status & $0, 1$ \\ 
relationship & $0, 1, 2, 3, 4$ \\ 
race & $0, 1, 2, 3, 4$ \\ 
sex & $0,1$ \\ 
capital-gain & $0,1$ \\ 
capital-loss & $0,1$ \\ 
hours-per-week & $1-99$ \\ 
country & $0,1$ \\ 
employment-type & $0, 1, 2, 3$ \\ 
salary & $0,1$ \\ 
\bottomrule
\end{tabular} 
\vspace{.5cm}
\label{tab:adult_features}
\end{table}

\paragraph{ELECTRICITY Dataset.}
The Electricity dataset \cite{harries1999splice}, represents energy pricing in Australia, over the period of May-1996 to December 1998, with recordings every 30 minutes giving 45312 samples. The dataset records the energy prices and demand for New South Wales and Victoria, and the amount of power transferred between the two states. The goal is to predict whether the transfer price increases or decreases.

The covariates outlined in Table \ref{tab:elect_features} are normalized to the interval $[0,1]$

We temporally partition the dataset where $\Dtrain$ is Mid-1996 to early-1997 and $\Dtest$ is early-1997 to 1998. We note that the dataset has been characterized as having concept shift for some features over the test period (however without an explicit timepoint or label). This could be due to behavioral changes or consumption pattern changes \cite{zliobaite2013good}. Hence, $\Dtest$ consists of data which is both congruous and incongruous with $\Dtrain$.

\begin{table}[h]
\centering
\caption{Summary of features for the ELECTRICITY Dataset \cite{harries1999splice}}
\begin{tabular}{ll}
\toprule
Feature & Range \\ 
\midrule
data & $0-1$ \\ 
period & $0-1$ \\ 
nswprice & $0-1$ \\ 
nswdemand & $0-1$ \\ 
vicprice  & $0-1$ \\ 
vicdemand & $0-1$ \\ 
transfer & $0-1$ \\ 
class & $0-1$ \\ 
\bottomrule
\end{tabular} 
\vspace{.5cm}
\label{tab:elect_features}
\end{table}

\subsubsection{Downstream Model}
In Section 4.3, we have a downstream task for each of the three datasets. For all the datasets, our base model which we train using $\Dtrain$ is a Random Forest Classification model with 100 estimators in the ensemble and splits are based on the 'Gini' criterion.
\clearpage
\section{Additional Experiments} \label{sec:appendixC}
This appendix presents additional experiments, validating further properties of Data-SUITE, conducting further comparisons or deep-dives into the regions identified by Data-SUITE and what insights can be garnered.

\subsection{Data-SUITE Ablation}\label{ablation}
Data-SUITE adopts a pipeline-based approach. Hence, to better understand the effect of each component, we perform an ablation study of different constituent components. The constituent components are then compared to the complete pipeline, which we denote as Data-SUITE (ALL).

The two components that we explicitly test are:
\begin{itemize}
    \item Data-SUITE (CONF): Test the conformal predictor without the representer. The conformal prediction process and instance stratification are as defined in the main paper.
    \item Data-SUITE (COP): Test the copula by itself. For this ablation, we fit the copula on $\Dtrain$. Then, to compute intervals per feature, we condition on the remaining features and sample 100 estimates from the copula. For example, for feature $x_1$, we condition on $x_2...x_n$. The uncertainty estimates are then the variance of these samples, which then is used to stratify the samples as before.
\end{itemize}

Firstly, we compare the coverage for each of the constituent components as per Table \ref{tab:ablation_coverage} for different configurations of the synthetic experiment. We see that Data-SUITE (All) is the only method to maintain coverage guarantees across all configurations. That said, we see a significant divergence between Data-SUITE (CONF) and Data-SUITE (COP), which suggests the conformal estimator is the most important component. The performance gap to Data-SUITE (All) is then likely the representer.
\begin{table}[h]
\centering
\caption{Coverage of constituent components of Data-SUITE}
  \scalebox{1}{
  \begin{tabular}{cl||ccc||cc}
    \hline
     & \multicolumn{1}{c||}{} & \multicolumn{3}{c||}{Proportion ($D_a$)} & \multicolumn{2}{c}{ Variance  ($D_b$)} \\ \hline
     \multirow{3}{*}{\STAB{\rotatebox[origin=c]{90}{}}}
      & PERTURBATION                     &  .1      &  .25     &   .5   & 1  &    2                           \\ \hline
     \multirow{9}{*}{\STAB{\rotatebox[origin=c]{90}{}}}
     &  \textcolor{ForestGreen}{\textbf{Data-SUITE (All)}}  & \textcolor{ForestGreen}{\textbf{.97}}  & \textcolor{ForestGreen}{\textbf{.96}} &  \textcolor{ForestGreen}{\textbf{.96}}  &  \textcolor{ForestGreen}{\textbf{.95}} & \textcolor{ForestGreen}{\textbf{.96}}   \\
     & Data-SUITE (CONF) &  .89 &  .88 &  .90  & .87  &  .90 \\
     & Data-SUITE (COP)   &  .16 & .15 &  .16  & .17  &   .16  \\ \hline
  \end{tabular}}
   \label{tab:ablation_coverage}
\end{table} 

Secondly, we compare the downstream MSE for each of the constituent components as per Table \ref{tab:ablation_mpi} for different configurations of the synthetic experiment. Recall that a lower MSE (i.e. closer to Train Data (BASELINE)) is desired. Data-SUITE (All) outperforms the constituent components and is less sensitive to perturbations. Interestingly, despite the higher coverage for Data-SUITE (CONF) vs Data-SUITE (COP), when evaluating for a downstream task, Data-SUITE (COP) in fact produces results which are less sensitive to perturbations. 
\begin{table}[h]
\centering
\caption{Downstream MSE for ablations of constituent components of Data-SUITE}
  \scalebox{1}{
  \begin{tabular}{cl||ccc||cc}
    \hline
     & \multicolumn{1}{c||}{} & \multicolumn{3}{c||}{Proportion ($D_a$)} & \multicolumn{2}{c}{ Variance  ($D_b$)} \\ \hline
     \multirow{3}{*}{\STAB{\rotatebox[origin=c]{90}{}}}
      & PERTURBATION                     &  .1      &  .25     &   .5   & 1  &    2                           \\ \hline
     & \textit{Train Data (BASELINE)}         &  \textit{.067}      &  \textit{.059}     &   \textit{.068}   &  \textit{.065}  &    \textit{.068}                   \\
     & Test Data            &     .222   &    .513   &   .889   &   .275      &   .889               \\ \hline\hline
     \multirow{9}{*}{\STAB{\rotatebox[origin=c]{90}{}}}
     &  \textcolor{ForestGreen}{\textbf{Data-SUITE (All)}}  & \textcolor{ForestGreen}{\textbf{.069}}  & \textcolor{ForestGreen}{\textbf{.122}} &  \textcolor{ForestGreen}{\textbf{.197}}  &  \textcolor{ForestGreen}{\textbf{.104}} & \textcolor{ForestGreen}{\textbf{.197}}   \\
     & Data-SUITE (CONF) &  .125 & .396 &  .846  & .293  &  .846  \\
     & Data-SUITE (COP)   &  .220 & .277 &  .451  & .236  &  .451  \\ \hline
  \end{tabular}}
   \label{tab:ablation_mpi}
\end{table} 

% We also assess the value of PCA as part of the pipeline, compared to an autoencoder. We reduce the representations to the same dimensionality for both. Replacing PCA by an Autoencoder, results in a DROP in MPI for all 3 datasets of 3\%, 6\% & 11\%. This motivates the choice of PCA.

\textbf{Takeaway:} Both results show reduced performance for the constituent components of Data-SUITE, with each component having different individual sensitivities. This highlights the necessity of the interconnected Data-SUITE (All) framework.
\newpage
\subsection{Regions identified by Data-SUITE as NOT OOD}\label{appendix:ood}
In this experiment, we address the question whether the regions identified by Data-SUITE as uncertain or inconsistent are in fact OOD.
We benchmark four widely used methods (with different detection mechanisms), which have been applied in the literature for OOD and outlier detection:
\begin{itemize}
    \item Mahalanobis distance \cite{lee2018simple}
    \item SUOD: Accelerating Large-\textbf{S}cale \textbf{U}nsupervised Heterogeneous
\textbf{O}utlier \textbf{D}etection \cite{zhao2021suod}
    \item COPOD: \textbf{Cop}ula-Based \textbf{O}utlier \textbf{D}etection \cite{li2020copod}
    \item Isolation Forest \cite{liu2012isolation}
\end{itemize}

We note for SUOD, much like the original paper, we make use of an ensemble of base estimators namely: Local outlier factor (LOF) \cite{breunig2000lof}, COPOD, Isolation Forest.

For each dataset, we have the instance IDs identified by Data-SUITE for various proportions. We then apply each of the aforementioned methods and compute the overlap between the predicted OOD/Outlier instances and our identified uncertain and inconsistent instances. 

\begin{figure}[!h]
\captionsetup[subfigure]{labelformat=empty, justification=centering}
\begin{subfigure}{0.33\textwidth}
\centering
\includegraphics[width=0.95\textwidth]{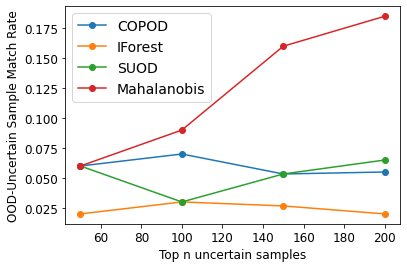}
\caption{SEER-CUTRACT}
\end{subfigure}%
\begin{subfigure}{0.33\textwidth}
\centering
\includegraphics[width=0.95\textwidth]{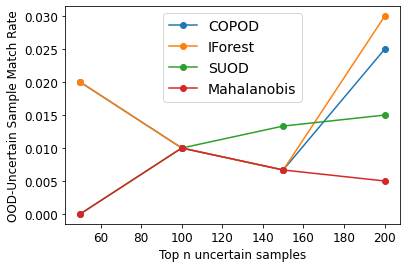}
\caption{ADULT}
\end{subfigure}%
\begin{subfigure}{0.33\textwidth}
\centering
\includegraphics[width=0.95\textwidth]{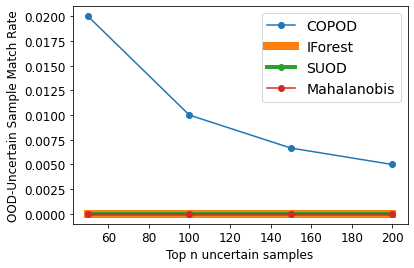}
\caption{ELECTRICITY}
\end{subfigure}%
\caption{OOD-Uncertain Instances Match Rate or overlap}
\label{fig:uncertain_ood}
\vspace{-0.5cm}
\end{figure}

The results for the overlap of uncertain instances can be found in Fig. \ref{fig:uncertain_ood}. We see minimal overlap across methods ranging between 2-18$\%$.

\begin{figure}[!h]
\captionsetup[subfigure]{labelformat=empty, justification=centering}
\begin{subfigure}{0.33\textwidth}
\centering
\includegraphics[width=0.95\textwidth]{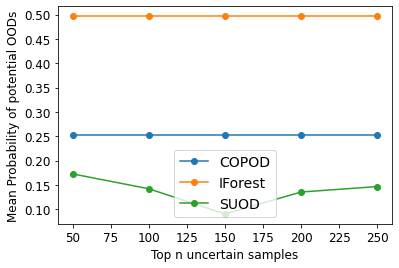}
\caption{SEER-CUTRACT }
\end{subfigure}%
\begin{subfigure}{0.33\textwidth}
\centering
\includegraphics[width=0.95\textwidth]{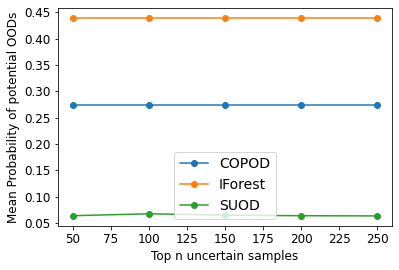}
\caption{ADULT}
\end{subfigure}%
\begin{subfigure}{0.33\textwidth}
\centering
\includegraphics[width=0.95\textwidth]{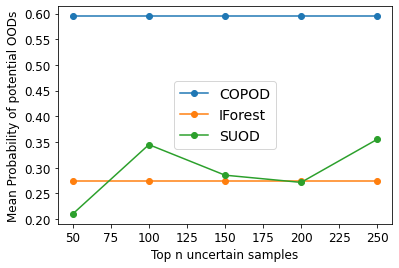}
\caption{ELECTRICITY}
\end{subfigure}%
\caption{Mean probability of predicted OOD, i.e. confidence}
\label{fig:proba_ood}
\vspace{-0.5cm}
\end{figure}

We additionally, evaluate the confidence scores (for those methods which provide confidence scores as outputs).  The goal is to see if an instance is predicted as OOD/outlier, then with what confidence does the detection method ascribe to the instance. The results in Fig. \ref{fig:proba_ood} suggest that the methods were often unconfident, with average confidence scores ranging between 5-50$\%$. This suggests the identified uncertain instances are unlikely OOD. 

\begin{figure}[!h]
\captionsetup[subfigure]{labelformat=empty, justification=centering}
\begin{subfigure}{0.33\textwidth}
\centering
\includegraphics[width=0.95\textwidth]{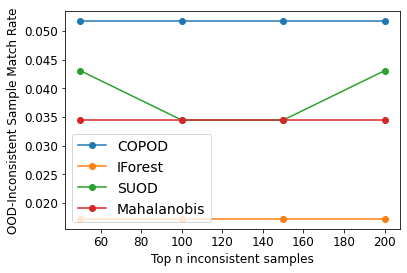}
\caption{SEER-CUTRACT}
\end{subfigure}%
\begin{subfigure}{0.33\textwidth}
\centering
\includegraphics[width=0.95\textwidth]{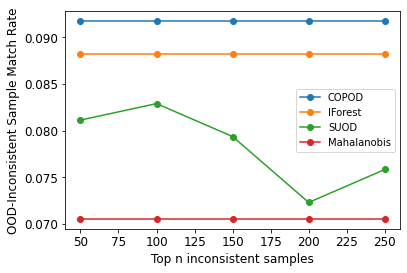}
\caption{ADULT}
\end{subfigure}%
\begin{subfigure}{0.33\textwidth}
\centering
\includegraphics[width=0.95\textwidth]{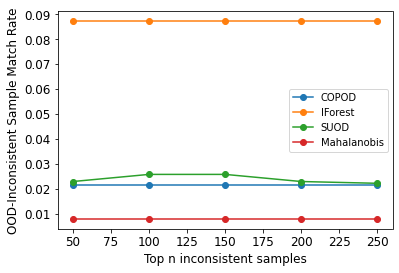}
\caption{ELECTRICITY}
\end{subfigure}%
\caption{OOD-Inconsistent Instances Match Rate or overlap}
\label{fig:incons_ood}
\vspace{-0.5cm}
\end{figure}

A similar question might be asked for the inconsistent instances. The results are similar to the uncertain instances as shown in Fig. \ref{fig:incons_ood}. This result, similar to uncertain instances, suggests the identified inconsistent instances are unlikely OOD.

\textbf{Takeaway:} The Data-SUITE identified \emph{uncertain} and \emph{inconsistent} instances are unlikely OOD. The reason is the limited overlap of predicted OOD and both \emph{uncertain} and \emph{inconsistent} instances, coupled with unconfident (low probability) predictions for OOD.

\subsection{EDA \& digestable prototype insights}\label{app:eda}
\begin{figure}[t]
\centering
\captionsetup[subfigure]{labelformat=empty, justification=centering}

\begin{subfigure}{1\textwidth}
\centering
\includegraphics[width=0.93\textwidth]{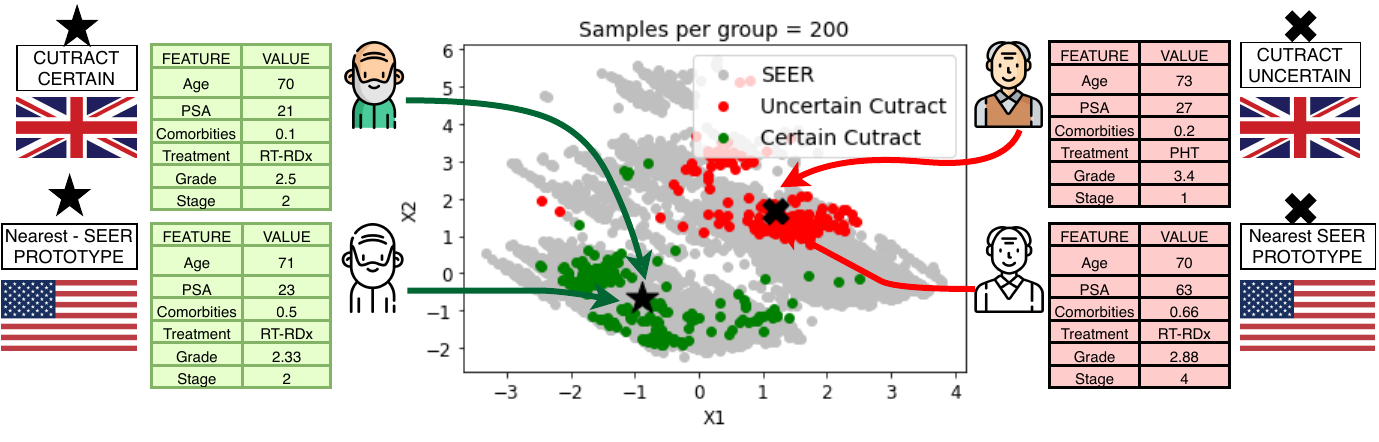}
\caption{i. SEER-CUTRACT: CUTRACT \emph{certain} instances are similar to their SEER nearest prototypes, whilst CUTRACT \emph{uncertain} instances are different to their nearest SEER prototypes (e.g. PSA). }
\end{subfigure}%
\\
\begin{subfigure}{1\textwidth}
\centering
\includegraphics[width=0.73\textwidth]{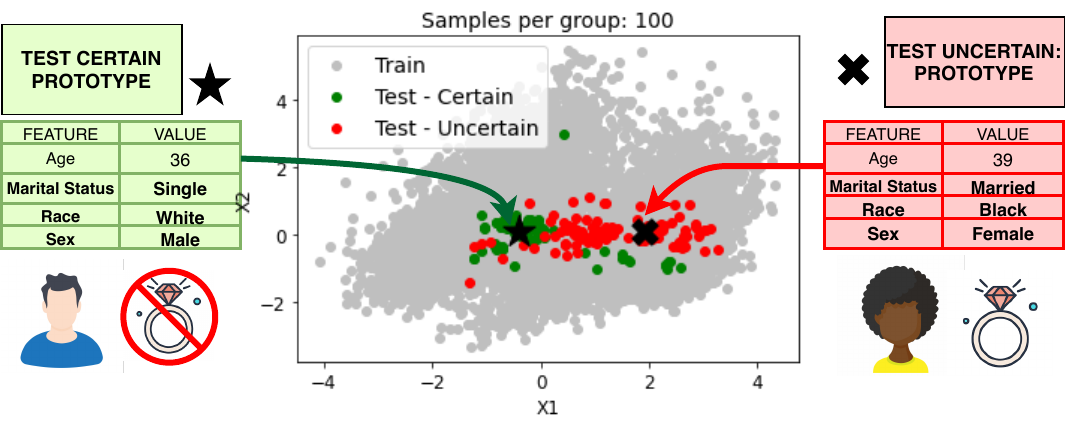}
\caption{ii. Adult: The \emph{certain} and \emph{uncertain} instances, represent two different demographics, aligning with the known dataset biases toward females. The \emph{uncertain} instances specifically highlight a sub-group of black, females.}
\end{subfigure}%
\\
\begin{subfigure}{1\textwidth}
\centering
\includegraphics[width=0.73\textwidth]{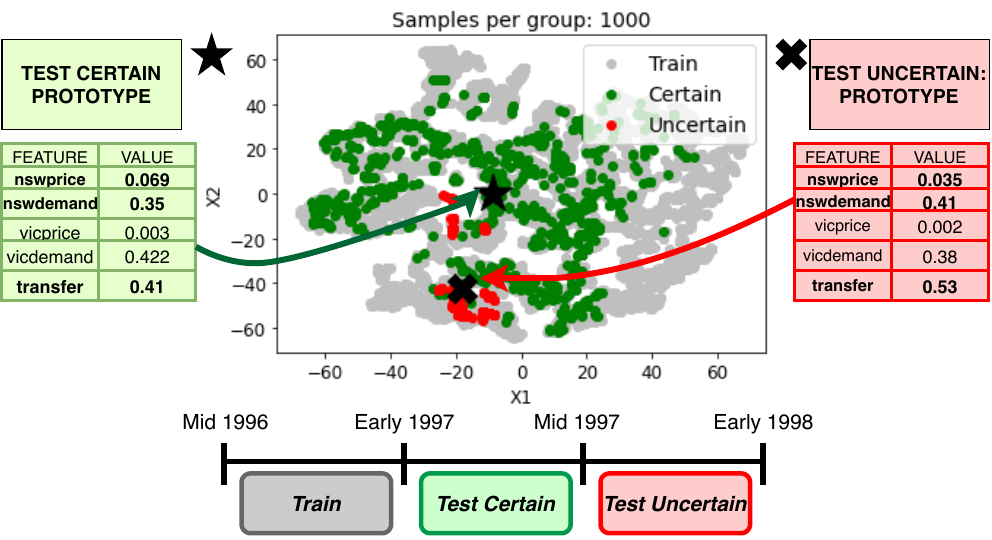}
\caption{iii. Electricity: The \emph{certain} instances are similar to the training set in features and time. The \emph{uncertain} instances identified, represent a later time period, wherein concept drift has likely occurred.}
\end{subfigure}%
\caption{Insights of prototypes identified by Data-SUITE. Tables describe the average prototypes for \emph{certain} and \emph{uncertain} instances.}
\label{fig:insights2}
\vspace{-0.5cm}
\end{figure}

As discussed in the main paper, we now conduct a more detailed analysis of the regions identified by Data-SUITE, as well as, the digestible average prototypes. This illustrates the full potential of the detail that practitioners can get from Data-SUITE and how they can practically use Data-SUITE to garner insights about the data. We present the diagram again for easy reference (see Fig \ref{fig:insights2}).

\textbf{\textit{SEER-CUTRACT.}} Data-SUITE identifies distinct \emph{certain} (Green) and \emph{uncertain} (Red) regions of ID data as shown in Fig. \ref{fig:insights2} (i). Beyond average prototypes for $\Dtest$ regions i.e., CUTRACT (UK), we also find the nearest neighbor SEER (USA) prototypes for each instance in the identified regions. Comparing these prototypes assists us to tease out the differences between the two geographic sites. We note three specific insights which can assist end-users. (1) \emph{certain} instances represent less severe patients than the \emph{uncertain} instances (see PSA values). (2) \emph{certain} instances:  CUTRACT (UK) and SEER (USA) prototypes are similar in all-feature values. (3) \emph{uncertain} instances: CUTRACT (UK) and SEER (USA) prototypes have differences in certain feature values (PSA, more comorbities, different treatment, staging score). 

We conduct an extended deep-dive below, highlighting features such as PSA, that show no difference on a population level, or for \emph{certain} instances. However, when teasing out \emph{uncertain} instances and their prototypes, we see this difference. This example illustrates the full potential of how Data-SUITE can be used by practitioners to uncover differences across sites (to benefit clinical practice), whilst also identifying patients where model performance would either be reliable or substandard (independent of the model).

\begin{figure}[h]
    \centering
    \includegraphics[width=0.45\textwidth]{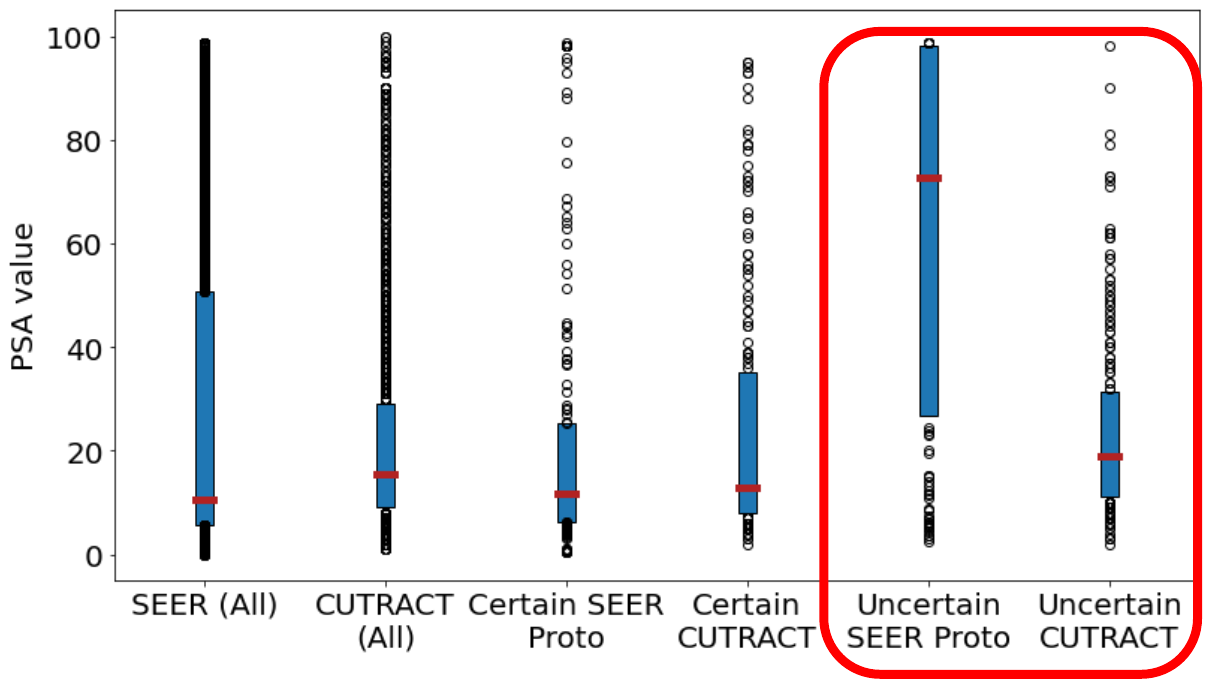}
    \caption{PSA Deep-Dive, which highlights how the difference is not evident at population level or for certain instances. Only the uncertain instances highlight the heterogeneity. However, it does highlight common support based on the whiskers of the box plot.}
    \label{fig:psa}
\end{figure}

We highlight this in Figure \ref{fig:psa}, which shows no PSA difference between the USA and UK on a population level, or for certain instances, but when teasing out \emph{uncertain} instances and their prototypes we see this difference. 

\begin{figure}[h]
\captionsetup[subfigure]{labelformat=empty, justification=centering}
\begin{subfigure}{1\textwidth}
\centering
\includegraphics[width=0.6\textwidth]{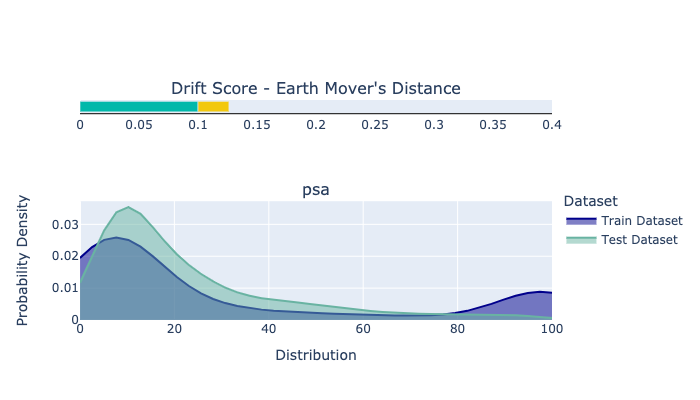}
\caption{Earth Movers Distance on the FULL Dataset (Population): Train (SEER) \& Test (CUTRACT) - No Difference}
\end{subfigure}%
\\
\begin{subfigure}{1\textwidth}
\centering
\includegraphics[width=0.6\textwidth]{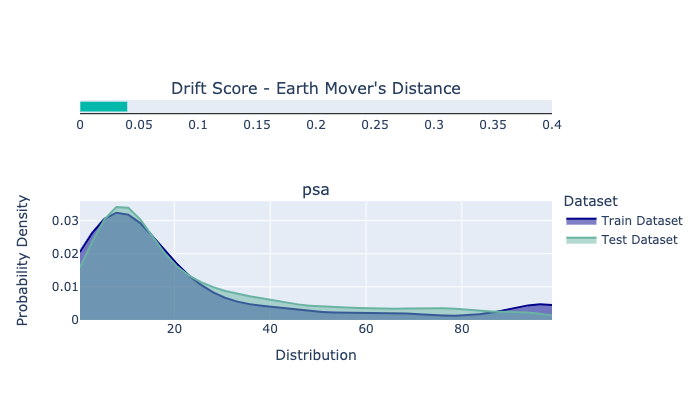}
\caption{Earth Movers Distance: \emph{CERTAIN}: Train (SEER Nearest Neighbors) \& Test (CUTRACT) - No Difference}
\end{subfigure}%
\\
\begin{subfigure}{1\textwidth}
\centering
\includegraphics[width=0.6\textwidth]{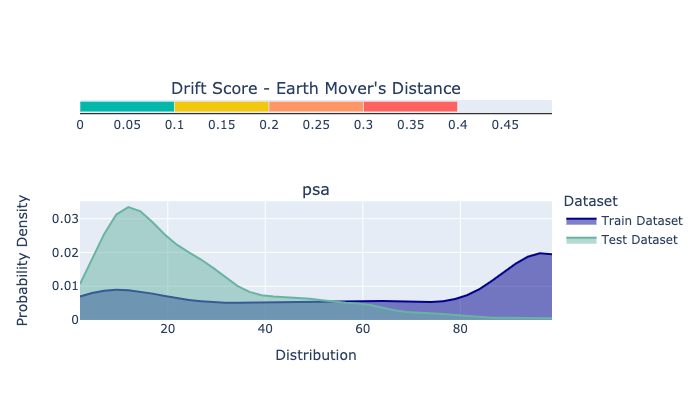}
\caption{Earth Movers Distance: \emph{UNCERTAIN}: Train (SEER Nearest Neighbors) \& Test (CUTRACT)- Clear Difference}
\end{subfigure}%
\caption{Earth Movers Distance Deep-Dive into PSA, which highlights how the difference is not evident at the population level or in certain instances. Only the uncertain instances highlight the heterogeneity }
\label{fig:emd}
\vspace{-0.5cm}
\end{figure}

We extend this analysis to further illustrate the full potential of the detail that practitioners can get from Data-SUITE. We do this by comparing the Earth Movers Distance (EMD) - see Fig. \ref{fig:emd} which is a common metric to flag drift. We see no PSA difference between the USA and UK on a population level, or for certain instances. However, when teasing out \emph{uncertain} instances and their prototypes, we see this difference. 

\textbf{Takeaway:} We highlight that while not evident at population level, the heterogeneous groups can be teased out by using Data-SUITE. This fully demonstrates the added capability of what Data-SUITE offers to practitioners.

\clearpage

\textbf{\textit{Adult.}} The dataset has a known bias between gender and income, which Data-SUITE successfully identifies. However, an added dimension between the most \emph{certain} and \emph{uncertain} instances as shown in Fig. \ref{fig:insights} (ii) is that marital status and race are also relevant biases. 

We capture this finding with prototypes, (i) \emph{certain} prototype: younger, single, white, males (ii) \emph{uncertain} prototype: older, married, black, females. 

These prototypes can inform end-users such as data scientists and stakeholders of the dangers of naively building models on this dataset without first considering these issues.

\textbf{\textit{Electricity.}} The dataset has concept drift over time, which Data-SUITE identifies as the \emph{uncertain} instances, which have increased transfer of energy between NSW and Victoria, increased NSW energy price and decreased demand relative to the \emph{certain} instances.

These prototypes can inform end-users of this change in consumption habits, despite the data still lying in-distribution. 

Additionally, we note as per Fig. \ref{fig:insights} (iii), the \emph{certain} instances capture the time-frame closer to the training data (Start 1997-Mid 1997), whilst the \emph{uncertain} instances are later in time (Mid 1997-Early 1998), hence more likely affected by the concept drift.

\subsection{Assessment of different CI regressors $g_i$}

In our formulation, we select a CI regressor ($g_i$). We wish to assess if the choice of CI regressor can impact the performance on a fixed downstream task predictive model. 

We hence carry out an experiment where we assess different $g_i \in \{RF, MLP, SVR and Tree\}$, based on the Mean Performance Improvement (MPI) -Eq. \ref{eq:perf_improvement} on a downstream task.

\textbf{Takeaway.} We find the variance of MPI across CI regressor types ranges between 0.04\%-3\% on our 3 datasets. The low variability in MPI is desirable, as it indicates minimal impact of the choice of CI regressor on Data-SUITE’s performance.

\subsection{Rank for a diverse set of downstream models}\label{rank_appendix}

\begin{figure}[!h]
\captionsetup[subfigure]{labelformat=empty, justification=centering}
\begin{subfigure}{0.45\textwidth}
\centering
\includegraphics[width=1\textwidth]{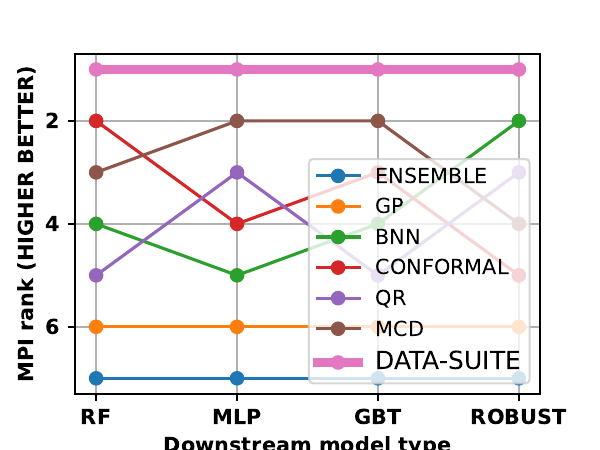}
\caption{SEER-CUTRACT}
\end{subfigure}%
\begin{subfigure}{0.5\textwidth}
\centering
\includegraphics[width=1\textwidth]{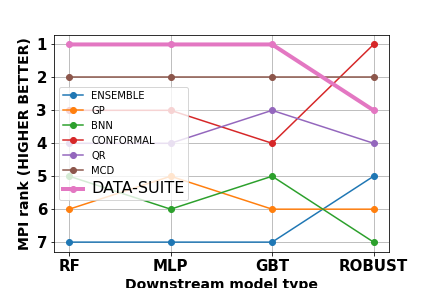}
\caption{ELECTRICITY}
\end{subfigure}%
\caption{Rank assessment for on diverse downstream models}
\label{fig:rank_mpi}
\end{figure}

We now present additional results for the experiment with a diverse set of downstream models, as outlined in Sec. \ref{diverse}. See Figure \ref{fig:rank_mpi}

\newpage
\subsection{Inconsistent instances $\lambda$ sensitivity}

As described in Section \ref{subsec:inconsistent_and_uncertain}, an instance $x$ is \emph{inconsistent} if the fraction of inconsistent features is above a predetermined threshold $\lambda \in [0,1]$: $\nu(x) > \lambda$.  In our implementation, we use $\lambda=0.5$, i.e. if more than 50\% of features are inconsistent, then the sample is considered inconsistent.

\begin{figure}[!h]
\captionsetup[subfigure]{labelformat=empty, justification=centering}
\begin{subfigure}{0.33\textwidth}
\centering
\includegraphics[width=1\textwidth]{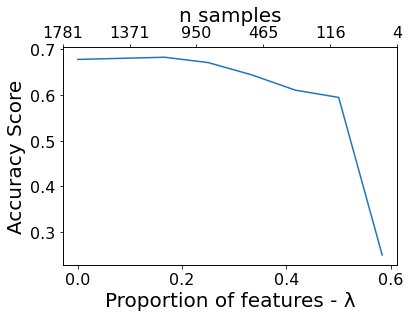}
\caption{SEER-CUTRACT}
\end{subfigure}%
\begin{subfigure}{0.33\textwidth}
\centering
\includegraphics[width=1\textwidth]{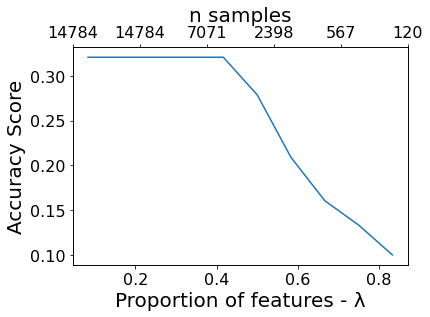}
\caption{ADULT}
\end{subfigure}%
\begin{subfigure}{0.33\textwidth}
\centering
\includegraphics[width=1\textwidth]{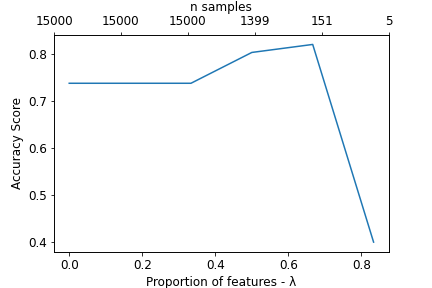}
\caption{ELECTRICITY}
\end{subfigure}%
\caption{$\lambda$ sweep to flag inconsistent instances}
\label{fig:lambda_sweep}
\vspace{-0cm}
\end{figure}

For completeness, we conduct an analysis of the sensitivity to the values of $\lambda \in [0,1]$. We show the accuracy score as a function of $\lambda$, as well as, the number of instances that would be classified as inconsistent for that value of $\lambda$. The results are shown in Fig. \ref{fig:lambda_sweep}. The flat, steady performance for small $\lambda$ and then drop off approximately around $\lambda=0.5$ across all datasets, suggests that our chosen value of $\lambda=0.5$ is indeed a sensible choice.

\textbf{Takeaway}: Based on the sweep across all three datasets, we see that $\lambda=0.5$ is a sensible choice in practice.

\subsection{Computation time comparison}
All experiments were run on CPU on a MacBook Pro with an Intel Core i5 and 16GB RAM. Besides the actual performance values, practitioners often are interested in the computation time associated with different methods. This is especially true if the algorithms are applied to very large datasets with many instances. 

We hence conduct a comparison of the computation time needed by each method to both train and to construct the predictive intervals for all instances in $\Dtest$. We present the results in Table \ref{tab:compute}, with the recorded computation time provided to the nearest minute.
\begin{table}[h]
    \centering
        \caption{Comparison of computation time across methods to the nearest minute}
    \scalebox{1}{
    \begin{tabular}{lccc}
\toprule
 &      SEER-CUTRACT & Adult & Electricity    \\
\midrule
\textbf{Data-SUITE}          &  4  & 1  & 1 \\
BNN & 2 &  1  &  1\\
CONFORMAL           & 2 &  1  & 2\\
ENSEMBLE            & 28 &  3  & 3\\
GP            & 22 &  15  & 24 \\
MCD            &  19 &  13  & 19 \\
QR & 1 &  1  &  1\\
\bottomrule
\end{tabular}}
    \label{tab:compute}
\end{table}

\textbf{Takeaway}: Data-SUITE does not have the fastest computation time (due to the inter-connected framework). However, the computation times are neither too dissimilar and nor prohibitive for practitioners to use.
\newpage
\subsection{Comparison to domain adaptation}
In the paradigm of domain adaptation, a popular metric is the maximum mean discrepancy (MMD) \cite{gretton2012kernel}.
For example, \cite{kumagai2019unsupervised,long2015learning,yanmmd,haeusser2017associative} have used MMD as the metric to compare distributions and subsequently, to optimize to minimize the distributional difference of representations.

The MMD metric is a distance-based metric that compares the mean embeddings of two probability distributions source $S$ and target $T$, in a reproducing kernel Hilbert space $H_k$. This is given by Eq. \ref{mmd}

\begin{align}\label{mmd}
    MMD(S,T) = ||\mu_{k}(S) - \mu_{k}(T) ||_{H_{k}}
\end{align}

We can compute unbiased estimates of samples from the two distributions after we apply the kernel trick (in this case we use a Radial Basis Function Kernel).

In the domain adaptation literature, the goal is to minimize the latent feature divergence by optimizing the MMD. Hence, MMD is a fundamental component of domain adaptation and is used to identify instances with discrepancies between source and target domains. 

As a comparison to our approach (Data-SUITE), we cast the problem as a potential domain adaptation problem and apply MMD, where the source distribution ($S \sim \Dtrain$) and target distribution ($T \sim \Dtest$). We then use the computed MMD to stratify instances, where lower MMD means \emph{certain} and large MMD means \emph{uncertain}. 

The mean performance improvement (MPI) is then computed for instances stratified based on MMD, similar to the experiment in Section \ref{experiment3}. Table \ref{tab:domain_adapt} contrasts the results using MMD vs Data-SUITE to stratify instances. 

\textbf{Takeaway.} The approach of Data-SUITE achieves greater average performance improvement based on MPI across all three datasets, when compared to framing the problem in the context of domain adaptation (using the MMD metric to identify and stratify instances).

\begin{table}[h]
    \centering
        \caption{Comparison of average performance improvement for Data-SUITE vs MMD (Domain Adaptation)}
    \scalebox{1}{
    \begin{tabular}{lcc}
\toprule
 &    \textbf{Data-SUITE}   & MMD    \\
\midrule
SEER-CUTRACT & \textbf{0.11} &  0.088   \\
Adult           & \textbf{0.64} &  -0.06   \\
Electricity            &   \textbf{0.26}  & -0.17 \\
\bottomrule
\end{tabular}}
    \label{tab:domain_adapt}
\end{table}

\subsection{Comparison to prototypes}
We compare the samples identified by Data-SUITE with prototypes. Specifically, we compare ProtoDash \cite{gurumoorthy2019efficient}.

\textbf{Takeaway.}  When explaining $\Dtest$ using prototypes, the prototypes uniformly cover the manifold, i.e. we cannot easily identify clusters of ``certai'' or ``uncertain'' as in Data-SUITE. 

Alternatively, explaining $\Dtest$ with prototypes from $\Dtrain$, the prototypes match the ``certai'' instances identified by Data-SUITE (DS), where the Pearson r $\approx$ 0.8 for DS(certain) vs prototypes. On the contrary, it doesn't match the uncertain samples where DS(uncertain) vs prototypes $\approx$ 0.55. 

Thus, the prototypes either are ``uniform samples'' or ``certain samples'' with no distinct ``uncertain samples''. We conclude that Data-Suite has an advantage of identifying BOTH ``certain'' and ``uncertain'' samples, relevant for reliable model deployment (Desiderata 2)

% \section{Old}
% \input{parts/old}

%%%%%%%%%%%%%%%%%%%%%%%%%%%%%%%%%%%%%%%%%%%%%%%%%%%%%%%%%%%%%%%%%%%%%%%%%%%%%%%
%%%%%%%%%%%%%%%%%%%%%%%%%%%%%%%%%%%%%%%%%%%%%%%%%%%%%%%%%%%%%%%%%%%%%%%%%%%%%%%

\end{document}